\documentclass{article}

\usepackage[T1]{fontenc}
\usepackage[utf8]{inputenc}
\usepackage{microtype}
\usepackage{inconsolata}
\usepackage{graphicx}
\usepackage{booktabs}
\usepackage{array}
\usepackage{hyperref}   %
\usepackage[preprint]{icml2026}
\usepackage{xurl}   %
\usepackage{amsmath,amssymb}
\usepackage{colortbl}
\usepackage{enumitem}
\usepackage{placeins}   %
\usepackage{fontawesome5}   %

\hypersetup{pdftitle={Manifold Pretraining Engineering: Right Geometry, Right Time, Right Place},
            pdfsubject={}, pdfauthor={Yihuai Hong, Hanxi Ma, Chen Zhao}}

\newcommand{\yh}[1]{\textcolor{purple}{#1}}
\definecolor{newpink}{RGB}{214,51,132}
\newcommand{\nw}[1]{\textcolor{newpink}{#1}}   %
\newcommand{\nwcolor}{\ifdraftmarks\color{newpink}\fi}   %
\definecolor{revorange}{RGB}{204,102,0}
\newcommand{\og}[1]{\textcolor{revorange}{#1}}   %
\definecolor{revcyan}{RGB}{0,140,170}
\newcommand{\cy}[1]{\textcolor{revcyan}{#1}}   %
\definecolor{revgreen}{RGB}{0,128,80}
\newcommand{\gn}[1]{\textcolor{revgreen}{#1}}   %
\definecolor{revblue}{RGB}{28,92,200}
\newcommand{\bl}[1]{\textcolor{revblue}{#1}}   %
\definecolor{revolive}{RGB}{125,125,0}
\newcommand{\fz}[1]{\textcolor{revolive}{#1}}   %
\definecolor{revbrown}{RGB}{150,90,30}
\newcommand{\sx}[1]{\textcolor{revbrown}{#1}}   %
\newif\ifdraftmarks
\draftmarksfalse
\ifdraftmarks\else
  \renewcommand{\yh}[1]{#1}\renewcommand{\nw}[1]{#1}\renewcommand{\og}[1]{#1}\renewcommand{\cy}[1]{#1}%
  \renewcommand{\gn}[1]{#1}\renewcommand{\bl}[1]{#1}\renewcommand{\fz}[1]{#1}\renewcommand{\sx}[1]{#1}%
\fi

\definecolor{gainbg}{HTML}{DCE9FA}
\definecolor{lossbg}{HTML}{FBE2D8}
\definecolor{headbg}{HTML}{F2F1ED}
\newcommand{\up}[1]{\cellcolor{gainbg}$\mathbf{#1}$}   %
\newcommand{\dn}[1]{\cellcolor{lossbg}$\mathbf{#1}$}   %

\newcommand{\cka}{\mathrm{CKA}}
\newcommand{\Lgeo}{\mathcal{L}_{\mathrm{geo}}}
\newcommand{\Ltask}{\mathcal{L}_{\mathrm{task}}}
\newcommand{\geo}[1]{\textsc{#1}}   %

\newcommand{\projicon}{\raisebox{-0.25em}{\includegraphics[height=1.15em]{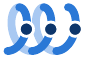}}}  %
\newcommand{\hficon}{\raisebox{-0.2em}{\includegraphics[height=1.05em]{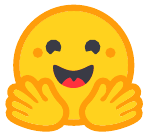}}}  %
\newcommand{\linkrow}{%
  \href{https://github.com/yihuaihong/manifolds}{\faGithub\ \texttt{Code}}\qquad
  \href{https://huggingface.co/YihuaiHong/manifold-pretraining-engineering}{\hficon\ \texttt{Models}}\qquad   %
  \href{https://yihuaihong.github.io/manifolds/}{\projicon\ \texttt{Project page}}}  %

\icmltitlerunning{Manifold Pretraining Engineering: Right Geometry, Right Time, Right Place}

\begin{document}

\twocolumn[
  \icmltitle{Manifold Pretraining Engineering:\\ Right Geometry, Right Time, Right Place}

  {\begin{center}   %
    {\bf Yihuai Hong}\textsuperscript{1,2}\quad
    {\bf Hanxi Ma}\textsuperscript{2,3}\quad
    {\bf Chen Zhao}\textsuperscript{2}\\[2pt]
    \textsuperscript{1}New York University\quad
    \textsuperscript{2}NYU Shanghai\quad
    \textsuperscript{3}Peking University\\[1pt]
    \texttt{yihuaihong@nyu.edu}\\[6pt]
    {\small \linkrow}   %
  \end{center}}

  \icmlkeywords{manifold priors, representation geometry, pretraining, interpretability, arithmetic}

  \vskip 0.2in
]

\makeatletter
\global\icml@noticeprintedtrue
\makeatother

\begin{abstract}
\sx{Interpretability has catalogued the manifolds on which language models represent concepts, such as a helix for numbers, but mostly to describe trained models. We turn them into a training technique, manifold pretraining engineering: an auxiliary loss that pulls each item's representation towards its own point on a target manifold.}   %
\sx{A helix prior raises the held-out addition accuracy of a 53M model pretrained on scarce arithmetic data from 18\% to 70\% (averaged over training, each model at its best learning rate); the gain holds at every size up to 1.07B parameters and, with the amount of data fixed, grows with model size.}   %
\sx{Six priors used together, five with targets that encode the answers, raise held-out accuracy, averaged over training, on all six concepts without detected interference; with them, a model reaches the final accuracy of the best model without priors after only 8--30\% of training. At 1.07B, the largest model we train, they raise accuracy on all six in every seed, by 49.9 points on numbers and 45.0 on dimensions, at no language-modelling cost on average.}   %
\sx{On an OLMo-2-7B checkpoint still learning addition, fine-tuning with the prior reaches 90\% held-out accuracy about five times sooner and ends at 100\%, against about 95\% without it.}   %
\sx{The prior helps most with the right geometry (a regular manifold matched to the computation), at the right time (while the task is being learned) and in the right place (the concept's own tokens, even in web text).}   %
\end{abstract}

\section{Introduction}
\label{sec:intro}

\begin{figure*}[t]
  \centering
  \includegraphics[width=\textwidth]{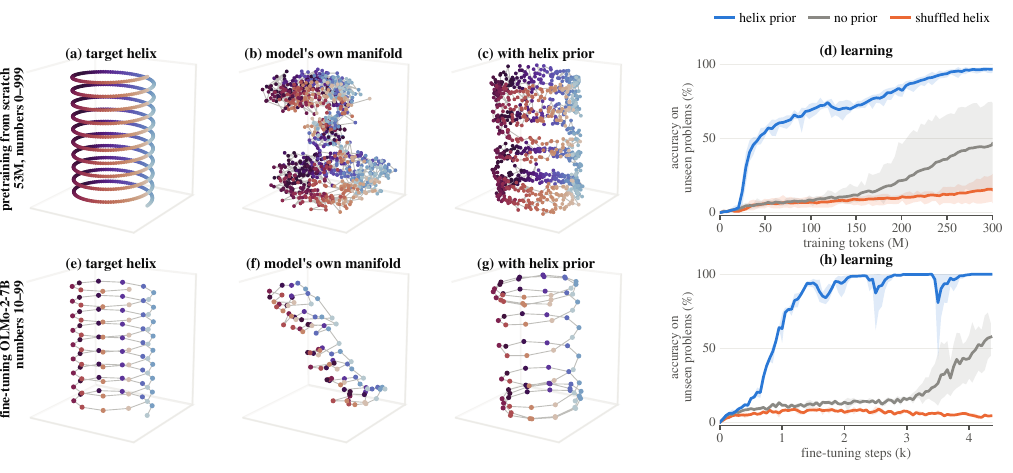}
 \caption{\og{A helix manifold prior shapes how a model represents numbers, and the model learns addition faster. Top: pretraining from scratch \fz{with 2\% arithmetic tokens, each arm except the shuffled helix at its own best learning rate.} Bottom: fine-tuning OLMo-2-7B at its 9B-token checkpoint on 7{,}000 addition pairs. (b,~c,~f,~g)~Representations, in axes computed with the help of the number labels (seed 0; CKA of the whole representation with the target: 0.62 in b, 0.94 in c). (d,~h)~Held-out accuracy (8 and 3 seeds; mean and min--max). Details: Appendix~\ref{app:capnotes}.}} %
  \label{fig:curves}
\end{figure*}

Interpretability research has found that language models organise many concepts on structured, low-dimensional manifolds \citep{gurnee2024spacetime,symmetry2026,goodfire_manifoldsteering}.
\nw{\og{They place numbers on a helix and on Fourier features used for addition \citep{zhou2024fourier,kantamneni2025trig}, and weekdays} and months on circles \citep{engels2025notlinear}.}
\fz{These geometries are part of how models compute: models add by manipulating their helix of numbers \citep{kantamneni2025trig}, and steering along a manifold rather than a single direction produces behaviour that the model shows naturally \citep{goodfire_manifoldsteering}.} %
\nw{We find that, for stepping tasks such as ``three months after March'', this geometry forms early in pretraining, long before the ability that uses it.} %
So far, interpretability has used this knowledge mostly to describe trained models. \sx{We ask whether it can also train them: \textbf{if a model needs a certain geometry to learn a task, can we supply that geometry during training so that the model learns the task faster?}}   %

\sx{We test this with a \emph{manifold prior}, an auxiliary loss that matches the relations among the model's representations of a concept's items to those of a target manifold, such as a helix for numbers.} By analogy with representation engineering \citep{zou2023repe}, we call the practice of designing such priors for training \emph{manifold pretraining engineering}: choosing which geometry to impose, when, and where.   %
\og{We do not search for the best target for each concept; \textbf{we ask whether, when and where a suitable one speeds up learning.}}   %

Arithmetic is our main testbed, because a model's capability on it is easy to measure and the right geometry is known.
\sx{We pretrain small language models from scratch with manifold priors, fine-tune intermediate checkpoints of OLMo-2 and Pythia (160M to 32B parameters) to vary how far a model has learned the task when the prior is introduced, and extend the priors to other operations, other concept domains and six priors at once (\S\ref{sec:window} and~\S\ref{sec:several}).}   %
\sx{A helix prior on numbers makes learning much faster (Figure~\ref{fig:curves}). In pretraining from scratch with scarce arithmetic data, it raises held-out addition accuracy, averaged over training, from 18\% to 70\% at 53M parameters. With 7\% arithmetic and 300M tokens at every size, it beats the model without a prior in all 8 paired seeds at each of nine sizes from 53M to 1.07B parameters, and the gain grows with model size (\S\ref{sec:faster}). In fine-tuning, a model that is still learning addition reaches 90\% held-out accuracy about five times sooner with the prior (\S\ref{sec:window}).}   %
Whether a prior helps depends on three conditions:
\begin{itemize}[leftmargin=*,itemsep=1pt,topsep=2pt]
 \item \textbf{Right geometry} (\S\ref{sec:faster} and~\S\ref{sec:geometry}). The manifold must be complete, regular and matched to the computation: the helix helps addition and hurts multiplication, and a shuffled helix \bl{with the same weight} hurts. \bl{A prior on numbers supplies the place-value structure that a whole-number tokenizer leaves the model to discover (\S\ref{sec:faster}), and it places numbers rather than passing information about answers (\S\ref{sec:geometry}).} %
 \item \textbf{Right time} (\S\ref{sec:time} and~\S\ref{sec:window}). \nw{The gain is largest while the model is learning the task and negative once it has learned it. \og{In pretraining, a prior in a window from 5\% to 40\% of training beats one from the first step, which also costs about eight times as much language-modelling quality as the window.}} %
 \item \textbf{Right place} (\S\ref{sec:place}). \nw{The prior acts on the concept's own tokens, even where numbers occur in ordinary web text. \gn{A model with the prior applied only there still beats the model without a prior, so the prior can be added to standard pretraining.}} %
\end{itemize}
\bl{Our contributions are:}
\begin{itemize}[leftmargin=*,itemsep=1pt,topsep=2pt]
 \item \bl{\textbf{A training technique.} Manifold priors, auxiliary losses based on linear CKA that pull the representations of a concept's items towards a target manifold, \gn{taken from interpretability or derived from the concept's structure. Unlike activation steering, which edits states at inference, or hidden-state distillation, which matches the states themselves, a prior constrains only the relations among a concept's items (\S\ref{sec:related}). The technique has four design choices: the target, the schedule, the site and the combination of several priors} (\S\ref{sec:method}).} %
 \item \bl{\textbf{Evidence in pretraining and fine-tuning.} With every arm at its own best learning rate, a helix prior on numbers raises held-out accuracy averaged over training at every model size we pretrained, \og{from 53M to 1.07B parameters (\S\ref{sec:faster}).} In fine-tuning checkpoints of OLMo-2 and Pythia, the gain follows the model's own capability and is largest while the task is being learned (\S\ref{sec:window}).}
 \item \sx{\textbf{Six priors at once.} Six priors trained together, five of them with targets that encode the answers, raise held-out accuracy, averaged over training, on all six of their concepts above the model without priors at its best learning rate for each (by 46.9 points on numbers), and interference between them is not detected. With them, a model reaches the final accuracy of the best model without priors after only 8--30\% of training, and with the priors kept on until the end it ends above that model on all six. Training them together also lifts dimensions by 13.5 points over its prior alone; this lift comes from the geometry of the torus and grid priors rather than their extra prior gradient, and inside the joint model, dimension answers use a code shared with the torus. At 1.07B, the largest model we train, the six priors raise accuracy on all six concepts in every seed, by 49.9 points on numbers, 49.2 on the grid, 45.0 on dimensions and 17.4 on the torus over the model without priors at its best learning rate, at no language-modelling cost on average. The gains do not shrink from 190M to 1.07B: the priors are higher in all 54 comparisons of sizes, seeds and concepts, and every one of the 21 models with them is above every one of the 63 models without them, at any learning rate and seed, on numbers, dimensions, the torus and the grid (\S\ref{sec:several}).}   %
 \item \bl{\textbf{The conditions and the limits.} The three conditions above, and what a prior does not do: \gn{\sx{it does not help when its geometry is imposed only at inference, without training}, and it does not fix a geometry the computation does not use. In a fine-tuning screen of 36 concept domains, among targets that do not encode the answer, the only robust gains were for numbers and months} (\S\ref{sec:limits}, Appendix~\ref{app:screen}).} %
\end{itemize}

\sx{These results point to another use of interpretability: the manifolds it finds can serve as training signals, not only as descriptions of trained models. Future work can test each newly found geometry as a prior, combine priors whose concepts share structure, and carry the approach to larger models and to concepts in ordinary text (\S\ref{sec:conclusion}).}   %

\section{\fz{Background and Related Work}}   %
\label{sec:related}

\paragraph{Concept manifolds in language models.}
\og{For numbers, language models add with Fourier features and a helix} \citep{zhou2024fourier,kantamneni2025trig}, \nw{alongside base-10 digit codes \citep{levy2025base10}. Different models learn similar number representations \citep{fu2026convergent},} and the geometry of number has been characterised more broadly \citep{numbergeometry}.
\nw{Beyond numbers, space and time are linearly decodable \citep{gurnee2024spacetime}, and weekdays and months lie on circles \citep{engels2025notlinear}. Such geometries can follow from symmetries in the statistics of language \citep{symmetry2026}, and models can reason about cyclic concepts with their base-10 addition \citep{feucht2026wild}.}
\sx{In small transformers trained from scratch on tasks over the same set of city coordinates, different tasks give different geometries of the world, training on several tasks makes them converge, and some tasks hinder the integration of new cities in later fine-tuning \citep{park2026convergent}.}   %
\nw{Mechanistic studies trace how models use these representations for arithmetic,} often through a bag of heuristics rather than one algorithm \citep{stolfo2023arith,nikankin2025heuristics}.
\nw{Representation geometry also changes during training \citep{tracinggeometry2025}, as does factual knowledge \citep{chang2024factual}.}
\sx{These studies analyse the geometry that models learn; we supply it during training and ask whether the model then learns the computation that uses it faster. In OLMo-2-7B, the geometry of numbers and months forms long before the ability to step through them (\S\ref{sec:window}), which leaves time for a prior to act.}   %

\paragraph{Steering and shaping representations.}
\nw{Activation steering mostly edits a single direction or contrastive vector \citep{arditi2024refusal,rimsky2024caa,zou2023repe}. Recent work instead argues} that local regions and manifolds are better objects of intervention, and that steering along a manifold yields behaviour the model shows naturally \citep{directions2regions,goodfire_manifoldsteering}.
At training time, internal states can be optimised directly, \fz{as in representation fine-tuning (ReFT; \citealp{wu2024reft})} and \nw{hidden-state distillation \citep{romero2015fitnets}. Aligning the features of a diffusion transformer} with those of a pretrained encoder makes it train much faster \citep{yu2025repa}. %
\nw{Our prior matches neither vectors nor hidden states. Instead, it uses linear CKA \citep{kornblith2019cka} to align the centred pairwise structure of a set of identified items with that of their target points\og{.}} \bl{Representation-similarity terms have also served as training regularisers \citep{li2019brains}; we use them to impose a target geometry, such as one that interpretability has found.}

\paragraph{Structural priors during training.}
In algorithmic tasks, generalisation can emerge long after memorisation \citep{power2022grokking}, \nw{through circuits whose progress can be measured \citep{nanda2023grokking}. \gn{Structural or geometric priors on the representation shorten this delay, and their effects depend on when they are applied} \citep{howe2026structure,howe2026windows,kazanskii2026tame}. More generally, deep networks have critical periods early in training, in which a temporary deficit has lasting effects \citep{achille2019critical}.}
An auxiliary objective that supplies the right intermediate quantity lets transformers learn multi-digit multiplication \citep{bai2025multiplication}.
For numbers in language models, \nw{the loss of a numerical probe has served as an auxiliary objective in fine-tuning \citep{yuchi2026numbers}, and post-hoc edits that make number representations more regular reduce arithmetic errors \citep{kadlcik2025numbers,stefanik2025universal}. Numerical structure can also be built into the input encoding \citep{zhou2025fone,mcleish2024abacus}.}
\og{We instead turn the geometries that interpretability has found into priors on the relations among a concept's items, and study when, where and with which geometry they help.} \bl{\gn{We reproduce in toy grokking how the effect of a prior depends on when it is applied (Appendix~\ref{app:toy})} and show that, in fine-tuning, \fz{the upper edge of the window in which a prior helps is set by} the model's own capability on the task and the lower edge mainly by the amount of training data (\S\ref{sec:window}).} %

\paragraph{\sx{Manifold constraints for optimization.}}   %
\sx{Another line of work also shapes geometry during pretraining, but for optimization: it constrains weight matrices to manifolds \citep{bernstein2025manifolds,an2026manifold}, normalizes weights and hidden states to the unit hypersphere \citep{loshchilov2025ngpt}, or projects the residual mixing matrices of hyper-connections onto the Birkhoff polytope of doubly stochastic matrices \citep{xie2026mhc}, while Muon-style optimizers orthogonalize the updates \citep{jordan2024muon,liu2025muon}. These are hard, task-agnostic constraints on scale, orthogonality or mixing, aimed at stable and fast training. Our priors are soft auxiliary losses on the relative geometry of selected representations (at one of the sites, the input-embedding rows of a concept's tokens), aimed at what the model learns about specific concepts.}

\section{Engineering Manifold Priors}
\label{sec:method}
\og{A manifold prior is an auxiliary loss that pulls a model's representations of a concept (such as numbers, weekdays or the elements of a group) towards a target manifold during training, each item towards its own point on it. We compare training with each prior against training without the prior (\geo{none}) and, in the main comparisons, also against a shuffled target, defined below.}
\fz{We judge a prior by how fast a model learns a task whose structure the target describes. The main task is addition, because the right geometry for numbers is known \sx{\citep{zhou2024fourier,kantamneni2025trig}}. In pretraining from scratch, addition problems make up a small share of the training text (\S\ref{sec:pretraining}); in fine-tuning, a pretrained checkpoint learns them from a small training set (\S\ref{sec:finetuning}). In both, we measure accuracy on held-out problems that never occur in training, averaged over training or as the time a model needs to reach 90\%. Other operations, concept domains such as months and weekdays, and, in pretraining, six concepts at once are tested in the same way (\S\ref{sec:several}, Appendices~\ref{app:details} and~\ref{app:domains}).} %

\paragraph{The loss.}
Let $X$ be the items of a concept, for example the numbers $0,\dots,999$ in pretraining, or $10,\dots,99$ in fine-tuning, and let $H_L\in\mathbb{R}^{n\times d}$ collect the residual-stream states of the $n$ items after layer $L$, where $d$ is the width of the model.
A target $G$ assigns each item $x\in X$ ideal coordinates $F_G(x)\in\mathbb{R}^{m}$, standardised per coordinate, and $F_G\in\mathbb{R}^{n\times m}$ stacks them for the $n$ items.
The prior is
\begin{equation}
  \Lgeo = \frac{1}{|\mathcal{S}|}\sum_{L\in\mathcal{S}}\bigl(1-\cka(H_L,F_G)\bigr),
  \label{eq:geo}
\end{equation}
added to the task loss $\Ltask$ (the cross-entropy of the answer, or the language-modelling loss in pretraining) with weight $\lambda$, where $\mathcal{S}$ is the set of layers at which the prior acts and $\cka$ is linear CKA \citep{kornblith2019cka},
\begin{equation}
  \cka(H,F)=\frac{\langle K_H,K_F\rangle_{\mathrm{F}}}{\lVert K_H\rVert_{\mathrm{F}}\,\lVert K_F\rVert_{\mathrm{F}}},
  \label{eq:cka}
\end{equation}
\nw{with the Frobenius inner product and norm, and with centred Gram matrices $K_H=CHH^{\top}C$ and $K_F=CF_GF_G^{\top}C$, where $C=I-\frac{1}{n}\mathbf{1}\mathbf{1}^{\top}$.}
Linear CKA is invariant \nw{to translation,} global rotations and isotropic scaling, so the loss is low only if the representation itself has the shape of the target \nw{up to these transformations}, \nw{with each item at its own target point. \og{In fine-tuning, a regression loss on a learned read-out would not constrain the shape (Appendix~\ref{app:loss}).}} %

\paragraph{Targets.}
For numbers the main target is the periodic code that interpretability work has found in pretrained models \citep{zhou2024fourier,kantamneni2025trig},
\begin{equation}
  F_{\textsc{helix}}(x)=\Bigl[\cos\tfrac{2\pi x}{T},\,\sin\tfrac{2\pi x}{T}\Bigr]_{T\in\mathcal{T}}\oplus\Bigl[\tfrac{x}{100}\Bigr],
  \label{eq:helix}
\end{equation}
where $\mathcal{T}=\{2,5,10,100\}$, $\oplus$ is concatenation, \nw{and the sine for $T{=}2$ is dropped because it vanishes on integers. For operands up to 999, we add the period 1000 and use $x/1000$ (\geo{helix3}).} %
\geo{digit} is a place-value code, $F_{\textsc{digit}}(x)=e_{\lfloor x/10\rfloor}\oplus e_{x \bmod 10}$, with $e_i$ a one-hot vector \citep{levy2025base10}.
\og{Three further targets separate the properties of the helix (Table~\ref{tab:geoms}). Since a Fourier spike at period $T$ is necessary but not sufficient for $x \bmod T$ to be linearly separable \citep{fu2026convergent}, \geo{helix\_nonsep} keeps the spikes without separability, \geo{sep\_ce} asks for separability alone, and \geo{helix\_coarse} has neither.} %
\nw{For a cyclic concept with $n$ items, the natural target follows from its symmetry: a circle, $[\cos\frac{2\pi v}{n},\sin\frac{2\pi v}{n}]$ for the item at position $v\in\{0,\dots,n-1\}$. For an ordered concept, it is a line.} %
\nw{In this way, we built targets for 36 concept domains, with circles for cycles, lines for orders, spheres for places and trees for taxonomies (Table~\ref{tab:domains36} in Appendix~\ref{app:screen}).} \sx{These hand-made shapes are coarse and may not fit every domain; we use them to test whether a fitting geometry speeds up learning, not to find the best shape for each domain.} \nw{Table~\ref{tab:helped} in Appendix~\ref{app:targets} collects the concepts for which a prior helped in our experiments.} %

\sx{A target can also be read out of a trained model: the states of the same items (e.g., the same numbers or colors) in a more mature model, or in a larger and stronger one, such as OLMo-2-7B for our 53M models. The two models can differ in width, because the prior compares only how the items relate to each other} (Appendix~\ref{app:mature}, \ref{app:pretraining}). %
\nw{Every target has a shuffled twin that assigns the same coordinates to a fixed permutation $\pi$ of the items, $F_{G\textsc{\_shuf}}(x)=F_G(\pi(x))$. It has the same dimensionality and loss scale, so it controls} for the strength of the auxiliary loss while breaking the correspondence between items and geometry.
\fz{Table~\ref{tab:geoms} in Appendix~\ref{app:targets} lists the number targets, and Table~\ref{tab:domains36} the targets of the 36 domains.} %

\paragraph{Schedule, site and multiple priors.}
The weight can depend on the training step $t$, and $K$ priors, for example one per concept, can be combined:
\begin{equation}
  \mathcal{L}=\Ltask+\sum_{k=1}^{K}\lambda_{k,t}\,\Lgeo^{(k)},
  \label{eq:general}
\end{equation}
with $\lambda_{k,t}=\lambda_k$ throughout, $\lambda_{k,t}=\lambda_k\,\mathbf{1}[w_0\le t/N<w_1]$ in a window of the $N$ training steps, or $\lambda_{k,t}=\lambda_k\,\mathbf{1}[\bar a_t<0.5]$ for a \emph{self-timed} prior that is on only while a running average $\bar a_t$ of the training-batch accuracy is below 50\%. %
With multiple priors we either divide one weight among them \og{($\lambda_k=\lambda/K$, \S\ref{sec:window}) or give each prior its own ($\lambda_k=1$, \S\ref{sec:several})}. %
The prior can act on separate prompts of the items (``\texttt{Q:~$x$}''), on their input-embedding rows, on the items where they occur inside the training data (\emph{in context}), or on the states that predict the answer. %

\paragraph{Fine-tuning setup.}
\og{We use LoRA \citep{hu2022lora} to fine-tune open intermediate checkpoints of OLMo-2 \citep{olmo2} (1B, 7B, 13B and 32B parameters, 9B--1678B training tokens) and of Pythia \citep{biderman2023pythia} (160M--12B parameters, 2B--286B tokens).} %
\nw{Unless stated otherwise, the prior uses $\lambda=20$ and acts at the layers at $1/8$, $1/4$, $3/8$ and $1/2$ of the depth.} %
The main task is two-digit addition, ``\texttt{Q:~$a$~+~$b$~=~}'' \nw{with $a,b\in[10,99]$. We train on 1{,}500 pairs, and \emph{accuracy} is the share of 1{,}000 disjoint test pairs for which the model's most likely next token is the sum.} %
\og{We also train on subtraction, a new operator $a\,\#\,b=(a+b)\bmod 100$, \fz{multiplication, a mixture of addition, subtraction and multiplication, and} five concept domains, and test transfer to unseen prompt formats (Appendix~\ref{app:details}).} %
We report the gain over training without the prior, $\Delta_G=\mathrm{acc}(G)-\mathrm{acc}(\textsc{none})$, \nw{in accuracy points with paired $t$-tests over 3 or 5 seeds. We call $\mathrm{acc}(\textsc{none})$ at a checkpoint the model's \emph{capability} at that checkpoint.} %

\paragraph{Pretraining setup.}
\sx{We pretrain OLMo-2-style models from scratch, from 53M up to 1.07B parameters and from 300M up to 1.2B tokens of FineWeb-Edu \citep{penedo2024fineweb}, with the OLMo-2 tokenizer; by default, a model has 53M parameters and trains on 300M tokens (Appendix~\ref{app:pretraining}).} %
\nw{In the main setting, every training sequence ends with one addition problem with operands up to 999 (2\% of the tokens). \fz{A setting with more arithmetic has four such problems (7\%).}} %
\nw{Held-out problems, which do not occur in training in either operand order, \og{\fz{are evaluated in training position, after about 500 tokens of text (Appendix~\ref{app:pretraining})}. \fz{We report held-out accuracy averaged over training and over the six combinations of operand lengths (the \emph{curve average}),} and the tokens needed to reach 90\% (median over 8 seeds).}} %
The prior (\geo{helix3}, $\lambda=5$, layers 2 and 4) acts in the window $[0.05,0.4)$ on separate ``\texttt{Q:~$x$}'' prompts, on the operands of the problems in context, or on numbers that occur in the web text. %
\nw{Main comparisons train each arm at its own best learning rate. \fz{In the helix-prior comparisons, both arms' learning rates are chosen on seeds 0--3 (0--1 at 1.07B),} and the comparison is repeated on four held-back seeds, whose held-out problems partly overlap those of the selection seeds (Appendix~\ref{app:pretraining}). \gn{Ablations use the learning rate $10^{-3}$, at which} the model without a prior is not tuned (Appendix~\ref{app:pt999}).} %
Earlier runs with operands 10--99 serve as finer ablations (Appendix~\ref{app:pretraining}). %

\paragraph{\bl{Why these settings.}}
\og{We pretrain from scratch to ask whether a prior speeds up learning a task when every arm is trained at its own best learning rate (\S\ref{sec:pretraining}), fine-tune checkpoints to vary how much of the task the model has already learned when the prior is applied and which geometry it is given (\S\ref{sec:finetuning}), use toy tasks to ask whether a prior can make a task learnable that no tuned training learns (\S\ref{sec:limits}), and screen 36 concept domains to ask how far the method reaches beyond numbers (Appendix~\ref{app:screen}).} \sx{In the main pretraining comparisons, each arm runs at its own best learning rate, so a gain cannot come from a rate that suits only one arm. A shuffled target of the same dimensionality and loss scale separates the geometry from the extra loss, and digit tokens, which supply place value through the input instead, are compared in \S\ref{sec:faster}.} %

\section{Manifold Pretraining}
\label{sec:pretraining}

\sx{We measure the speed-up at each model's best learning rate and size (\S\ref{sec:faster}), find when and where the prior helps most (\S\ref{sec:time} and \S\ref{sec:place}), and train six priors at once (\S\ref{sec:several}).}   %

\subsection{Faster Learning at Each Model's Best Learning Rate}
\label{sec:faster}

\paragraph{Scarce arithmetic data.}
\nw{In the main setting, with 2\% arithmetic tokens, the helix prior on the operands in the window $[0.05,0.4)$ raises held-out accuracy averaged over training from 18.4\% to 69.9\% at 53M parameters, from 29.1\% to 78.5\% at 190M and from 24.4\% to 77.4\% at 505M (means over 8 seeds). \og{At each size the curve average is higher with the prior in all eight paired comparisons, including the four held-back seeds, and the best run without the prior is below the worst run with it} (Figure~\ref{fig:curves}d; Figure~\ref{fig:scaling}a in Appendix~\ref{app:pttpp}).} %
\nw{\og{Without the prior, 0 of 8 seeds reach 90\% within 300M tokens at 53M, 2 of 8 at 190M, near the end of training, \fz{and 0 of 8 at 505M at the selected learning rate; with the prior, 8 of 8 reach it at 505M.}} \og{With four times as many tokens, every seed without the prior reaches this level at 53M, but the prior keeps a lead of 16.5 points in curve average (85.1\% against 68.7\%, 8 of 8 seeds).}} %
\nw{\gn{In these longer runs, the model with the prior first reaches 90\% 2.1 times sooner, stays above it 1.1--1.2 times sooner and reaches 50\% 3.6 times sooner.} The prior helps most early in learning.} %

\paragraph{More arithmetic data.}
\sx{With 7\% arithmetic tokens every model learns the task, and the prior changes how fast: at each arm's best learning rate it reaches 90\% 3.7 and 3.9 times sooner at 53M and 190M (first crossing, over all 8 seeds) and 4.2 times sooner at 505M (on the selection seeds), and at 1.07B, with learning rates chosen on two seeds, it first reaches 50\% 4.7 times sooner on the four held-back seeds (Tables~\ref{tab:pt999} and~\ref{tab:pt7lr} in Appendix~\ref{app:pt999}). Its gains in curve average at nine sizes are given below.} %

\begin{figure}[t]
  \centering
  \includegraphics[width=\columnwidth]{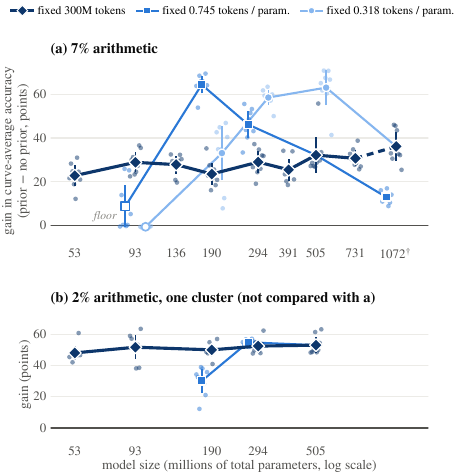}
  \caption{\og{How the gain changes with model size depends on what is held fixed. Gain of the helix prior in curve-average held-out accuracy (mean of 8 paired seeds with 95\% interval; dots: seeds), each arm at its own learning rate. Tokens per parameter count non-embedding parameters. Hollow markers are at the floor and excluded from the trend tests (pre-registered rule). \sx{At 300M tokens, the solid line joins the eight sizes run on L40S GPUs; $^{\dagger}$on B200 GPUs.} Details: Appendix~\ref{app:capnotes}.}}
  \label{fig:tpp}
\end{figure}

\paragraph{\nw{Gain and model size.}}
\nw{How the gain changes with model size depends on what is held fixed (Figure~\ref{fig:tpp}).} %
\og{We also pretrain models from 93M to 1.07B parameters with 7\% arithmetic at two fixed ratios of tokens to non-embedding parameters, 0.318 and 0.745, each arm at its own learning rate (Appendix~\ref{app:pttpp}).} %
\sx{With the number of tokens fixed, the gain increases with model size. At 300M tokens with 7\% arithmetic, the helix prior raises held-out accuracy averaged over training by 22.8 to 36.1 points at each of nine sizes from 53M to 1.07B (ten cells of size and GPU type, 8 of 8 seeds in each; Figure~\ref{fig:tpp}, Table~\ref{tab:pt300}). On the eight sizes run on L40S GPUs, a pre-registered test finds the gain increasing with size, by 1.66 points per $e$-fold of non-embedding parameters (0.96 with median and 1.19 with trimmed-mean gains; seed-bootstrap 90\% interval $[+0.71,+2.68]$, positive on both halves of the seeds; 90\% interval from the spread of the sizes $[+0.03,+3.30]$, two-sided $p=0.096$).} %
\sx{The three sizes not seen before the test was registered do not show the increase on their own (mean slope $+0.93$, median $-0.48$, trimmed $+0.41$; seed-bootstrap 90\% interval $[-0.71,+2.68]$), so they show only that the new sizes did not fall below the earlier trend. The gain is not monotonic (it dips at 190M and 391M). Along this ladder, the number of tokens per non-embedding parameter falls from 21 to 0.5, and the increase comes from the model without the prior getting weaker ($-1.55$ points per $e$-fold; median $-0.83$, trimmed $-1.11$), while the model with the prior stays at 89--90\% (Appendix~\ref{app:pttpp}).} %

\nw{With the ratio of tokens to non-embedding parameters fixed, so that larger models also see more data, \nw{there is no consistent trend. \og{At 0.745 the gain falls from 64.4 points at 190M to 13.0 at 1.07B, a fall that the pre-registered test finds only by excluding the 93M model, which is at the floor.} \fz{At 0.318 it rises from 33.1 at 190M to 63.0 at 505M and falls back to 36.1 at 1.07B.}}} %
\og{The two results fit together because the gain varies mainly with the number of training tokens: within one size it first rises and then falls (at 190M: 33.1, 64.4 and 23.5 points at 36M, 84M and 300M tokens), peaking roughly between 70M and 130M tokens with 7\% arithmetic, and with more tokens the model without a prior catches up (at 1.07B and 703M tokens, 79.3\% against 92.3\% with the prior). At a fixed ratio the larger models train on more tokens and lie further down the falling side (Appendix~\ref{app:pttpp}). The number of tokens varies together with the amount of arithmetic seen, the timing of the window and the prior's share of the gradient, and this design does not separate these factors.} %

\begin{figure*}[t]
  \centering
  \includegraphics[width=\textwidth]{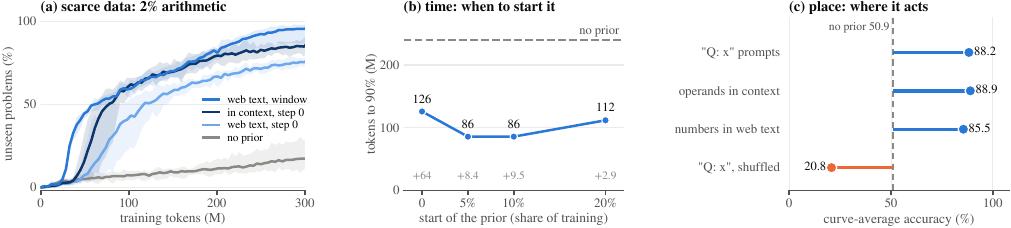}
 \caption{\og{Manifold pretraining from scratch (53M parameters, 300M tokens, learning rate $10^{-3}$). (a)~2\% arithmetic, 8 seeds (mean and min--max); ``step 0'': the prior is on from the first step; ``window'': in $[0.05,0.4)$ of training. (b)~7\%, 4 seeds: \fz{the prior on numbers in web text, in a window ending at 40\% of training}; grey: increase in validation bits per byte ($\times10^{-3}$). (c)~7\%, 4 seeds: curve-average accuracy by where the windowed prior acts. Details: Appendix~\ref{app:capnotes}.}} %
  \label{fig:pretraining}
\end{figure*}

\paragraph{\fz{The geometry, not just an auxiliary loss.}} %
\og{In pretraining, we ran the shuffled helix (\S\ref{sec:method}) only at 53M parameters.}
With 2\% arithmetic, in the same window and for the same steps, it gives no gain ($-1.8$ points on the operands and $-0.9$ in web text, \nw{against $+62.0$ and $+56.1$ for the correct helix).} %
With 7\% arithmetic it lowers the curve average by 23.6 points on the operands. %
\gn{The gain therefore requires the correct mapping from values to the geometry, and a wrong geometry can lower the curve average.}

\paragraph{What the helix supplies.}
\og{Because every number up to 999 is one OLMo-2 token, the model must learn that 47 is 4 tens and 7 units.} %
\og{Digit tokens alone make the model reach 90\% 6.6 times sooner (operands 10--99, learning rate $10^{-3}$), about as fast as with the best prior on whole-number tokens, and with digit tokens the prior adds little.} %
\og{Other place-value targets also help: one circle per digit almost as much as the helix (84.9\% vs.\ 88.2\%) and, with operands 10--99, the number geometry read out of OLMo-2-7B, a noisy helix or a units-plus-tens code 75--85\% as much as the helix; targets that encode only the number of digits do not.} %
\gn{A prior on numbers supplies the place-value structure that a whole-number tokenizer leaves the model to discover.}

\subsection{\bl{Right Time:} A Window from 5\% to 40\% of Training}
\label{sec:time}

\paragraph{Not from the first step.}   %
\og{With the web-text prior in a window ending at 40\% of training,} \nw{the model reaches 90\% after 86M tokens when the prior starts at 5\% or 10\%, after 112M when it starts at 20\%, and only after 126M when it starts at the first step (Figure~\ref{fig:pretraining}b).} %
\fz{\og{A prior adds little before the model has formed any number representation, and the later it starts after that, the smaller the gain.}} %
\og{Applied from the first step to the end of training, the same prior also falls behind: with 2\% arithmetic, the window beats it by 18.0 points of curve-average accuracy, at 0.035 bits per byte less (Figure~\ref{fig:pretraining}a).} %

\paragraph{A fixed window is enough.}
\og{A self-timed prior that also starts at 5\% of training but switches off once training accuracy reaches 50\% \fz{is 1.6--2.0 points below the window on sums of two three-digit numbers in each of four settings (7\% arithmetic, 53M)}, with no detectable difference in language-modelling cost, so the pre-registered decision rule selects the fixed window, which needs no task signal.} %

\paragraph{The right time also lowers the cost.}
\og{A prior from the first step raises the final validation bits per byte by 0.017--0.057, whether it stays on or is self-timed, mostly in the first thousand steps; a shuffled target costs as much as the correct one.} %
\nw{\gn{A window starting at 5\% \fz{costs about eight times less than the same window started at the first step} (0.008 vs.\ 0.064 bits per byte for the prior on web text).} With every model at its own best learning rate, the cost is \og{0.6--0.9\% of the validation bits per byte} after 300M tokens and undetectable after 1.2B.} %

\subsection{\bl{Right Place:} The Concept's Own Tokens}
\label{sec:place}

\paragraph{On the numbers, wherever they occur.}
The prior is about equally effective on the operand tokens of the problems in the training batch and on separate ``\texttt{Q:~$x$}'' prompts \nw{of the numbers: in the window, curve-average accuracy is 88.9\% and 88.2\%} (7\% arithmetic, 4 seeds; Figure~\ref{fig:pretraining}c). %
On numbers in web text it is \og{slower at our default weight,} \gn{but there the ratio of its gradient to the language-modelling gradient is also the largest of the three sites}, \fz{and with self-timed priors from the first step, the gap to the operands shrinks from 7.3 to 2.5 points at $\lambda=1$}. %
\og{With operands 10--99, \fz{the prior applied throughout reached 90\% sooner on the operands than on the prompts (after 27M vs.\ 70M tokens)}, and on the token that predicts the answer, a helix of the sum hurts (curve average 31.9\% vs.\ 63.9\%).} %
\gn{For the gain, the numbers themselves must take the correct geometry; the prior need not act where the sum is computed.}

\paragraph{In ordinary text: a pretraining plug-in.}
\og{A prior on operands needs the positions of the problems, known only for synthetic data. Applied instead only to numbers that occur naturally in web text, and never to the problems, the prior still speeds up addition, so it can be added to standard pretraining.} %
\og{With 2\% arithmetic and at the operand prior's learning rate, \fz{it gains 13.0 points of curve-average accuracy at 53M with 1.2B tokens and 43.5 points at 190M} (8-seed means, 8 of 8 positive).} %

\subsection{Multiple Manifolds at Once}
\label{sec:several}

\sx{We pretrain the 53M model with six priors at once: the helix for numbers, a lattice for physical dimensions ($\{-1,0,1\}^3$), a sphere for cities and one for countries (``which of two is closer to a third?''), a torus for the group $\mathbb{Z}_6\times\mathbb{Z}_6$ and a grid for sums and differences on $\{-2,\dots,2\}^2$; dimensions, the torus and the grid are written as made-up tokens.} %
\sx{Every sequence carries the 2\% addition of the main setting and two problems from the other domains. All arms of a seed see identical data and differ only in which priors are on and in learning rate, and each prior acts in the window on its items in context with $\lambda=1$. The six priors run at $10^{-3}$, a rate chosen on separate seeds and not tuned for this set; the model without priors runs at six learning rates from $2\times10^{-3}$ to $6.4\times10^{-2}$ (the highest added post hoc). The pre-registered tests use 8 fresh seeds (Appendix~\ref{app:domains}).}   %

\newcommand{\tabci}[1]{{\scriptsize\color{black!60}#1}}   %
\begin{table}[t]
  \centering\small
  \setlength{\tabcolsep}{3pt}
  \resizebox{\columnwidth}{!}{\begin{tabular}{@{}lrrr@{\hspace{3pt}}lrrr@{\hspace{3pt}}l@{}}
    \toprule
     & \multicolumn{2}{c}{Accuracy (\%)} & \multicolumn{4}{c}{Six $-$ none} & \multicolumn{2}{c}{Six $-$ one} \\
    \cmidrule(lr){2-3}\cmidrule(lr){4-7}\cmidrule(l){8-9}
    Manifold & None & Six & Gain & \tabci{95\% CI} & End & Faster & Lift & \tabci{95\% CI} \\
    \midrule
    Numbers & 23.6 & 70.5 & $+46.9$ & \tabci{[42.4, 51.5]} & $+33.9$ & 3.9$\times$ & $-0.5$ & \tabci{[$-$1.1, \phantom{0}0.1]} \\
    Dimensions & 47.3 & 81.2 & $+33.9$ & \tabci{[26.3, 41.6]} & $+30.7$ & 4.4$\times$ & $+13.5$ & \tabci{[\phantom{$-$}7.4, 19.6]} \\
    Cities & 86.3 & 92.6 & $+6.3$ & \tabci{[\phantom{0}2.7, 10.0]} & $+3.7$ & 2.6$\times$ & $+1.7$ & \tabci{[$-$0.3, \phantom{0}3.7]} \\
    Countries & 89.3 & 95.1 & $+5.8$ & \tabci{[\phantom{0}3.4, \phantom{0}8.1]} & $+3.7$ & 2.1$\times$ & $+1.2$ & \tabci{[\phantom{$-$}0.0, \phantom{0}2.4]} \\
    Torus & 63.0 & 85.0 & $+22.0$ & \tabci{[15.2, 28.8]} & $-3.0$ & 4.1$\times$ & $+4.4$ & \tabci{[\phantom{$-$}0.6, \phantom{0}8.2]} \\
    Grid & 41.0 & 60.6 & $+19.7$ & \tabci{[12.6, 26.7]} & $-1.6$ & 5.0$\times$ & $+7.4$ & \tabci{[$-$2.0, 16.7]} \\
    \bottomrule
  \end{tabular}}
  \caption{\sx{Six priors pretrained together (53M, 8 fresh seeds): curve-average held-out accuracy (\%) without priors, at the best of six learning rates for that manifold, and with the six (at $10^{-3}$); the paired gain with its 95\% interval over seeds (positive in 8 of 8 seeds on every manifold), the gain at the end of training (End) and how many times sooner the six get 90\% of the way from chance to the final accuracy without priors (Faster). Six $-$ one: the six priors minus that manifold's prior alone, both at $2\times10^{-3}$ (95\% interval). Dimensions: the strict held-out set. Except for numbers, the targets encode the answers. Details: Appendix~\ref{app:domains}.}}   %
  \label{tab:comb}
\end{table}

\paragraph{\sx{All six manifolds gain together.}}

\sx{With the six priors, every manifold is learned better and sooner. In 8 of 8 seeds, each of the six is above the model without priors at every one of the six learning rates we tried, from $2\times10^{-3}$ to $6.4\times10^{-2}$ (7 of 8 for the grid at the highest rate). Against the best of these rates, which lies inside the range for every manifold, the curve average rises by 46.9 points on numbers, 33.9 on dimensions, 6.3 on cities, 5.8 on countries, 22.0 on the torus and 19.7 on the grid (Holm $p\le0.0023$; Table~\ref{tab:comb}). The model reaches the end-of-training accuracy of that best model after only 8--30\% of training, on every manifold and in every seed, and gets 90\% of the way there from chance 2.1 to 5.0 times sooner. Interference is not detected: at $2\times10^{-3}$, joint training costs at most 0.5 points on any manifold relative to its prior alone, and every 95\% interval excludes a loss larger than half of that prior's gain (Appendix~\ref{app:domains}). Language modelling still dominates training: at $10^{-3}$ its loss carries 71\% of the gradient norm during the window, and the final validation loss, 1.213 bits per byte, lies between those of the two best models without priors (1.210 at $2\times10^{-3}$ and 1.215 at $4\times10^{-3}$).}   %

\paragraph{\sx{Joint training lifts dimensions.}}
\sx{Training the priors together also helps dimensions beyond its own prior. At $2\times10^{-3}$, the six lift dimensions by 13.5 points over the dimensions prior alone (8 of 8 seeds; Holm $p=0.0037$; a direction we did not pre-register). The gap widens after the priors switch off, because the prior alone then loses more accuracy on dimensions (32.9 against 12.4 points, 8 of 8 seeds); at the end of training the six lead it by 25.0 points on dimensions (7 of 8 seeds, one tie) and 11.7 on the torus (8 of 8). The lift needs the geometry of the torus and grid priors, which are structurally similar to dimensions: without these priors it is only 1.5 points (12.0 less, 7 of 8 seeds), and with their targets shuffled at the same strength, the lift disappears (8 of 8 seeds). Inside the joint model, dimension answers use a code shared with the torus: removing the torus answer subspace cuts held-out dimensions accuracy by 44--46 points with the six priors, against 1.5--5 with the dimensions prior alone, more than control subspaces of the same removed variance do, and the coupling runs both ways. Most of the lift does not need later practice on the torus and the grid: with their problems removed from training as the window closes, 77\% of it remains (7 of 8 seeds), and 82\% in a rerun that keeps every other problem in its position (8 of 8 seeds).}   %

\paragraph{\sx{Kept on until the end.}}
\sx{Kept on until the end of training, the six priors give the highest final accuracy of all our runs on all six manifolds. In every seed, they end above the model without priors at its best rate by 31.4, 35.2, 2.5, 2.1, 4.1 and 33.2 points on numbers, dimensions, cities, countries, the torus and the grid, and above the run switched off after 40\% of training by 12.3 points on dimensions, 9.4 on the torus and 38.5 on the grid, at a final validation loss only 0.1\% above the lowest without priors. Training longer does not replace the joint priors on dimensions. The dimensions prior alone reaches their level on the strict held-out dimension questions while it is on, but then its dimension geometry fades (CKA with the target from 0.96--0.98 to 0.75--0.79), and it ends at 44.5 and 29.5\% after twice and four times the training, against 77.3\% for the six priors at the original length. The model without priors, at its best rate for dimensions, reaches that level only after four times the training (78.1\%), and the six kept on until the end stay far above it: 96.5 against 78.1\% on the strict dimension questions and 91.4 against 50.8\% on the grid, each of the eight seeds above both of its seeds.}   %

\sx{For the five domains other than numbers the target encodes the answers, so these gains show that the model uses structure it is given; this run has no shuffled target, and the six tests share the same runs. Several reads are post hoc: the growth of the lift after the window, the runs kept on until the end, and their comparison with four times the training, which spans two batches. The longer runs have two seeds per cell, one learning rate per arm and a cosine schedule stretched over the longer run, and the six priors were not trained longer. The reads of the shared code use two re-drawn pairs of runs and edit only the answer position, and the share of the lift carried by practice, about a quarter, is still growing at the end of training (Appendix~\ref{app:domains}).}   %

\paragraph{\sx{The gains do not shrink up to 1.07B.}}   %
\sx{At 1.07B, the largest model we train, the six priors raise accuracy averaged over training on all six concepts in every seed, by 49.9 points on numbers, 49.2 on the grid, 45.0 on dimensions and 17.4 on the torus, each against the model without priors at its best learning rate, with the learning-rate range closed on both sides. The gains do not shrink with model size: from 190M to 1.07B they stay at 50--51 points on numbers (51.3, 51.4, 49.9), 43--48 on dimensions (48.1, 42.8, 45.0), 37--49 on the grid (42.9, 37.1, 49.2) and 17--20 on the torus (20.4, 18.2, 17.4). Every one of the 21 models with the six priors, in the window or kept on until the end, is above every one of the 63 models without priors, at any learning rate and seed, on numbers, dimensions, the torus and the grid: averaged over training, at least 75.3, 80.0, 86.0 and 68.5\% against at most 32.3, 64.6, 76.5 and 49.1\%. In all 54 comparisons (3 sizes $\times$ 3 seeds $\times$ 6 concepts) the model with priors is higher, and in every seed it also ends higher on numbers, dimensions, the torus and the grid. At 1.07B this comes at no language-modelling cost: at the same learning rate the bits per byte are equal on average (0.0\%; higher in 3 of 4 seeds), while at 190M and 505M the six cost a little ($+0.3\%$ and $+1.4\%$, 4 of 4 seeds each), and at no size does any of five zero-shot benchmarks move by more than 0.6 points. At 190M, the speed-up and the joint lift seen at 53M hold in all three seeds: the model reaches the best final accuracy without priors after 8--36\% of training on numbers, dimensions, the torus and the grid, and the lift on dimensions is $+10.4$ points; at 1.07B it is preliminarily positive ($+6.6$; $+5.9$ at the end, 3 of 3 seeds), and at 505M it does not hold ($-1.5$). These runs keep the 53M recipe; at 1.07B, a one-seed check puts the learning rate of the six-prior runs on a plateau (Appendix~\ref{app:domains}).}   %

\paragraph{\sx{Where the priors come from.}}
\sx{The city and country priors passed a pre-registered screen of 24 concept domains, one prior at a time, and replicated on 8 fresh seeds, in tests that share their comparison runs; a screen of six small algebraic structures on three seeds picked the grid and the torus (Appendix~\ref{app:domains}).}   %

\section{\bl{Fine-Tuning Checkpoints: Time and Geometry}}
\label{sec:finetuning}

\subsection{The Prior Helps While the Model Is Learning}
\label{sec:window}

\paragraph{Representations come before abilities.}
\og{In OLMo-2-7B, \fz{the geometry of numbers, number words, ordinals, months and hours reaches half} its final strength by 21B tokens, but the ability to step through them (``three months after March'') does so only after 84B or 630B tokens, depending on the domain (Appendix~\ref{app:atlas}).} %
\nw{\gn{A prior can act after the geometry has formed and before the ability has.}}

\paragraph{Faster learning.}
At the 9B-token checkpoint of OLMo-2-7B, fine-tuned on 7{,}000 pairs, \og{the model with the prior reaches 50\% held-out accuracy after 900 steps (4{,}150 without it) and 90\% after 1{,}400, and ends at 99.9\% after 10 epochs, against 61.9\% without it} (Figure~\ref{fig:curves}h). \og{The earlier the checkpoint, the larger and the longer-lasting the advantage: at 21B tokens, the model without a prior catches up within 10 epochs (Appendix~\ref{app:details}).} %
\sx{Given enough steps, task-only fine-tuning comes close, reaching 90\% about five times later and ending at about 95\% after 30 epochs; a shuffled helix with the same weight does not reach 50\% within 10 epochs (Appendix~\ref{app:details}).}   %
\sx{The gain holds with LoRA, ReFT and full fine-tuning and on a held-out prompt format, and general ability is unchanged} (HellaSwag, ARC and LAMBADA within 0.7 points of the base model; \citealp{zellers2019hellaswag,clark2018arc,paperno2016lambada}; Appendix~\ref{app:details}, \ref{app:tables}).   %

\begin{figure*}[t]
  \centering
  \includegraphics[width=\textwidth]{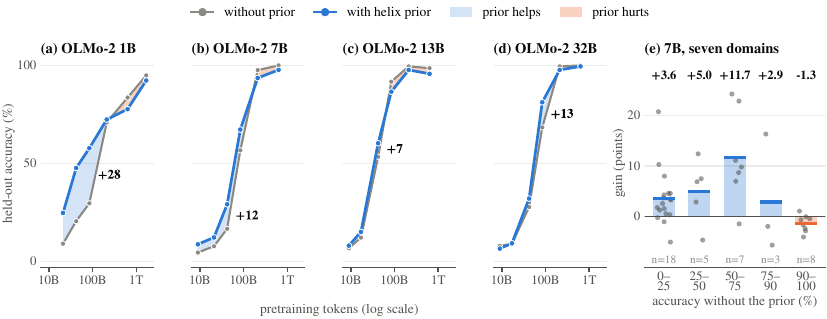}
 \caption{\nw{The benefit window. (a)--(d)~Held-out accuracy of OLMo-2 1B--32B on two-digit addition at each pretraining checkpoint, without and with the helix prior; the shaded gap is the gain, and the number marks the largest gain. (e)~OLMo-2-7B on seven domains at six checkpoints: mean gain per bin of capability (the accuracy without the prior), with the individual domain--checkpoint points. 5-seed means where available, otherwise 3 seeds. Details: Appendix~\ref{app:capnotes}; Pythia: Figure~\ref{fig:pythia}.}} %
  \label{fig:window}
\end{figure*}

\paragraph{\fz{Capability sets the benefit window.}} %
\og{For every OLMo-2 size (Figure~\ref{fig:window}a--d), the prior helps while task-only accuracy is rising and stops helping once it saturates: the largest gains are $+6.9$ to $+28.1$ points, and for all four sizes the gain is negative once capability exceeds about 80\%.} %
Pythia shows the same pattern \og{(Figure~\ref{fig:pythia})}, and so do subtraction and a new operator \og{$a\,\#\,b=(a+b)\bmod100$ (Appendix~\ref{app:details}).} %
Across seven domains of OLMo-2-7B (Figure~\ref{fig:window}e), \og{the mean gain peaks at $+11.7$ points between 50 and 75\% capability and is $-1.3$ above 90\%.} %
\og{The gain is thus an inverted-U function of the model's own capability: the prior speeds up a computation the model is forming but does not create one it is not forming.}

\paragraph{The lower edge is set by data.}
\sx{At the earliest checkpoints the gain is small with 1{,}500 training pairs but $+70.4$ points at 9B and 21B tokens with 7{,}000, where the task-only model does not even fit its training set within three epochs; the gain is specific to the geometry and grows with the number of pairs at every checkpoint (Appendix~\ref{app:tables}).}   %
The lower edge of the window \nw{thus reflects mainly the training budget, not the maturity of the number representation;} the upper edge reflects the model's capability. %

\paragraph{\sx{Timing on its own.}}
\og{A self-timed prior, active only while training accuracy is below 50\%, can be used without knowing where the window is. It keeps the full gain inside the window ($+11.1$ vs.\ $+11.2$ points at 42B tokens), removes almost all of the harm after it ($-0.5$ vs.\ $-3.9$ at 210B) and switches itself off sooner the more capable the model is (Appendix~\ref{app:details}).} %

\subsection{\bl{Right Geometry: Complete, Regular and Matched to the Computation}}
\label{sec:geometry}

\begin{table}[t]
  \centering\small
  \setlength{\tabcolsep}{5pt}
  \begin{tabular}{@{}lrrr@{}}
    \toprule
    \multicolumn{4}{@{}l}{\emph{(a) Components of the helix}} \\
    \addlinespace[1pt]
    & \multicolumn{1}{c}{1B} & \multicolumn{2}{c}{7B} \\
    \cmidrule(lr){2-2}\cmidrule(l){3-4}
    Target & 42B & 42B & 84B \\
    \midrule
    \geo{helix}         & \up{+29.5} & \up{+10.8} & $+9.9$ \\
    \geo{sep\_ce}       & $+7.3$           & $+4.1$           & \up{+4.2} \\
    \geo{digit}         & \up{+10.0} & $+3.6$           & $+2.8$ \\
    \geo{helix\_coarse} & $-1.5$           & $-3.5$           & $-3.7$ \\
    \geo{helix\_nonsep} & $-2.4$           & \dn{-5.4}  & $-7.7$ \\
    \midrule
    \multicolumn{4}{@{}l}{\emph{(b) One 7B model trained on three operations}} \\
    \addlinespace[1pt]
    Tokens & \multicolumn{1}{c}{$a+b$} & \multicolumn{1}{c}{$a-b$} & \multicolumn{1}{c}{$a\times b$} \\
    \midrule
    42B  & \up{+13.0} & \up{+18.9} & \dn{-11.8} \\
    84B  & \up{+12.3} & \up{+20.5} & \dn{-11.2} \\
    210B & $-3.5$           & $-1.5$           & \dn{-8.8} \\
    \bottomrule
  \end{tabular}
 \caption{Gain over \geo{none} in accuracy points (3 seeds); blue: significant gain, orange: significant loss ($p<0.05$, paired \nw{$t$-test).}
  (a)~Targets that isolate parts of the helix (Table~\ref{tab:geoms}), OLMo-2 1B and 7B at the given checkpoints (training tokens).
  (b)~\geo{helix} on OLMo-2-7B fine-tuned on a mixture of addition, subtraction and multiplication; each column is that operation's own test set.
 }
  \label{tab:geometry}
\end{table}

\paragraph{The whole helix, and a regular one.}
\nw{Table~\ref{tab:geometry}a takes the helix apart. A target with its Fourier spikes but no linear separability of $x \bmod T$ (\geo{helix\_nonsep}) gives no gain in any of three settings, while separability alone (\geo{sep\_ce}) \fz{gives 25--42\% of the helix's gain}, about as much as the digit code. The order \geo{helix} $>$ \geo{sep\_ce} $\approx$ \geo{digit} $>$ \geo{none} $>$ \geo{helix\_coarse} $\ge$ \geo{helix\_nonsep} is the same in all three settings. \gn{These targets test by intervention the account that \citet{fu2026convergent} built from correlations.}} %
\sx{Noise added to the helix lowers the gain monotonically in both settings we tested, and targets less regular than the model's own number geometry give a negative gain: the prior makes the model's existing, roughly helical manifold more regular (Appendix~\ref{app:targets}).}   %

\paragraph{Matched to the computation, within the operation.}
Addition and subtraction are translations in the helix coordinates; multiplication is not.
\og{When the model is fine-tuned on all three operations, one shared helix raises addition and subtraction by 12--21 points and lowers multiplication by 11--12 (Table~\ref{tab:geometry}b).} %
\sx{Applied only to the second operand of addition and subtraction problems, the helix raises both more than a shared helix in the same experiment and leaves multiplication almost unharmed, and a shuffled helix there gives no gain (Appendix~\ref{app:opcond}).}   %

\paragraph{\bl{Inputs and outputs.}}
When 15 operand values never occur in fine-tuning, most of the gain on problems that contain them remains even if the prior never sees these \og{values ($+8.4$ of $+13.1$ points at 42B tokens; Appendix~\ref{app:further}),} so the prior places numbers rather than passing information about answers. \bl{\gn{In short, for addition the prior helps on the operands. \og{In pretraining, a helix of the sum on the token that predicts the answer hurts addition (\S\ref{sec:place}), whereas} in fine-tuning on weekdays and months a circle on that token helps (Appendix~\ref{app:resample}).}} %

\section{What a Prior Does Not Do}
\label{sec:limits}

\paragraph{It supplies values; it does not teach them.}
\sx{We looked for tasks that a model learns with a prior but not in any tuned training without one (criteria in Appendix~\ref{app:walls}). The closest case is three-digit numbers held out of every arithmetic problem, so that they appear only in web text: without a prior, those from 700 to 999 are added correctly at most about 11\% of the time in every configuration we tested, and a helix prior on their embedding rows, kept on to the end, raises this to 99\%. The prior supplies the values rather than the model finding them: given the coordinates of another number, the model answers 94--99\% of these problems as if it were that number (7 of 8 seeds), and once the prior stops maintaining the coordinates, they are overwritten and accuracy falls. On toy tasks where tuned training stays at the floor, a prior partly raises held-out accuracy but makes none of them learnable in the settings we ran.}   %

\paragraph{\nw{Given the structure, a model learns to use it.}}
\sx{Only one case, a toy, meets our criteria, with a target that encodes the whole table. A one-layer transformer trained full-batch on 6.7\% of the sums and differences of the 125 vectors of $\{-2,\dots,2\}^3$ learns them in 0 of 5 fresh seeds in each of 28 AdamW settings without a prior, with four times the budget (settings with weight decay below 0.1 were added after the pre-registered test had passed). With a prior towards the coordinates of the vectors, all 5 seeds reach the pre-registered 95\% held-out accuracy at some evaluation (one only intermittently), at the lowest of the three learning rates we ran with it, and a random and a shuffled target do not. The wall comes from this small sample of a large table, not from the operation, and the case is one cell of a screen (Appendix~\ref{app:walls}).}   %

\paragraph{It does not fix a geometry the computation does not use.}
\og{No target we tried, whether the logarithm, prime exponents or per-place codes, helps multiplication in the training format (Appendix~\ref{app:mult}).} %

\paragraph{\sx{It does not help when imposed only at inference.}}   %
\og{Snapping representations onto the ideal manifold at inference never helps (at three OLMo-2 sizes), whereas replacing the same coordinates with those of a wrong number collapses accuracy wherever the model can add, so the model reads its operands from this subspace (Appendix~\ref{app:further}).} %

\paragraph{\nw{What it costs.}}
\nw{Table~\ref{tab:costs} in Appendix~\ref{app:walls} collects the costs we measured.} %
\sx{The larger costs in the table come from using a prior outside the conditions of \S\ref{sec:intro}, at the wrong time or with the wrong geometry.} %

\section{Conclusion and Future Work}
\label{sec:conclusion}

We turned \nw{the manifolds that interpretability research has found in language models} into a training technique, manifold pretraining engineering.
\gn{A manifold prior makes models learn faster in fine-tuning and, with each model at its own best learning rate, in pretraining from scratch at every model size we tested, from 53M to 1.07B parameters. With scarce data, it determines whether a model learns a task within the budget. It leaves general ability unchanged in fine-tuning,} \fz{costs 0.6--0.9\% of the validation bits per byte after 300M tokens of pretraining} \nw{and no detectable amount after 1.2B tokens at 53M parameters}, and acts on learning rather than at inference (Appendix~\ref{app:further}). %
\sx{Priors can also be combined. Six used together, five of them with targets that encode the answers, raise accuracy, averaged over training, on all six of their concepts, and interference between them is not detected. With them, the model reaches the final accuracy of the best model without priors after only 8--30\% of training; kept on until the end of training, they end above the best model without priors on all six. At 1.07B, the largest model we train, they raise accuracy on all six concepts in every seed, by 49.9 points on numbers and 45.0 on dimensions, and the gains do not shrink from 190M to 1.07B (\S\ref{sec:several}).}   %
The results give a recipe: \fz{derive the target from the concept's structure or, in pretraining, read it out of a trained model;} apply it to the concept's own tokens, for example to numbers wherever they occur in ordinary text; switch it on at about 5\% of training rather than at the first step\bl{, and in fine-tuning while the model is still learning the task, for example with a self-timed prior}; \sx{when priors are combined, keep them on until the end of training and keep the language-modelling loss dominant}; and expect the largest gains where the tokenizer leaves structure for the model to discover. %
\og{Used outside these conditions (from the first step, with a shuffled geometry, on a computation the geometry does not match, on a model that has already learned the task or, for addition in pretraining, as a helix of the sum on the token that predicts the answer), the same prior helps less, not at all, or hurts.}

\og{The next steps} are larger models, longer training and more concepts in ordinary text. %
Beyond numbers, the atlas of decodable manifolds (Appendix~\ref{app:atlas}) offers many candidate targets, but in a fine-tuning screen of 36 domains the geometry itself helped only numbers and months (Appendix~\ref{app:screen}).\sx{ In pretraining, priors for countries and cities, whose targets encode the answers, pass a screen of 24 domains, replicate on fresh seeds and gain again in the set of six (\S\ref{sec:several}).} %
\sx{Inside the joint model of six priors, dimension answers use a code shared with the torus (\S\ref{sec:several}); if such shared codes cause the joint lift, choosing priors whose domains share a manifold becomes a principle for combining them.}   %
\yh{We invite others to design finer targets by hand, to derive them automatically and efficiently, or to read stronger ones out of trained models, and our code includes all 36 targets and the screen as a starting point. Finding the geometry that a computation needs, and when, is the open problem.} %

\section*{Limitations}

\nw{Our tasks are synthetic (arithmetic on numbers up to 999 and simple concept arithmetic in fixed prompt formats), and arithmetic} is the only skill for which we show the full technique.
\sx{Pretraining uses small models (53M to 1.07B parameters) trained on at most 1.2B tokens, at most about 23 tokens per parameter, far below the hundreds to thousands of current LLMs, with only the learning rate tuned, and fine-tuning uses intermediate checkpoints rather than finished models. Whether gains in final accuracy hold at such ratios is open: on numbers, at a fixed model size, the gain shrinks once training is long enough for the model without a prior to catch up, and in fine-tuning the prior helps only while a model is still learning the concept, whereas on dimensions and the grid the six priors kept on until the end stay above the model without priors trained four times as long (\S\ref{sec:several}).}   %
\sx{All our runs use AdamW; whether the priors carry over to optimizers that constrain weights or orthogonalize updates (\S\ref{sec:related}) is open.}   %
How much faster the prior makes learning depends on the tokenizer (much less with digit tokens) and on how speed is measured \nw{(\S\ref{sec:faster})}. %
The shuffled control places a wrong geometry rather than none, so it shows that the correct mapping from values to geometry is needed, \gn{not how much gain a neutral regulariser would give.} %
\sx{With the six priors of \S\ref{sec:several} switched off after 40\% of training, the torus and the grid end training slightly below the best models without priors (the torus by 3.0 and 1.1 points at the two best rates, the grid by 1.6); kept on until the end, all six end above them (post hoc).}   %
\sx{Above 53M, the six priors run with three seeds per size, and the learning rate of the six-prior runs was checked only at 1.07B, with one seed. At the same learning rate, the six priors cost a little language-modelling loss at 190M and 505M (0.3 and 1.4\% bits per byte, 4 of 4 seeds each), and none on average at 1.07B.}   %
In fine-tuning, the weekday and month tests measure transfer across equivalent shifts\nw{. With a strict split, the month gain remains, but the weekday test is at the floor (Appendix~\ref{app:details})}. %
Targets are hand-designed or read out of trained models, and linear CKA sees only the linearly accessible part of a representation. \bl{\gn{To use the prior, one must also identify the items of a concept, as tokens or prompts, and, for the in-context and web-text sites, where they occur in the training data.}}
\sx{The main open question is whether the conditions we find, a regular target on the concept's own tokens in a window early in training, carry over to pretraining on trillions of tokens and to skills less clearly structured than arithmetic.}   %

\section*{Impact Statement}

\fz{This paper presents work whose goal is to advance the field of machine learning by using what interpretability has learned about the representations of language models to train them. Manifold priors make models learn some structured tasks faster; we study them in models of up to 32B parameters and on tasks such as arithmetic, which carry no direct risk. We do not see consequences that need to be highlighted beyond those of making language models more capable or cheaper to train in general.} %

\bibliography{references}
\bibliographystyle{icml2026}

\appendix
\section{Experimental Details}
\label{app:details}

\paragraph{Checkpoints.}
\og{Table~\ref{tab:ckpts} lists the OLMo-2 stage-1 checkpoints; their Hugging Face revisions are named \texttt{stage1-step$S$-tokens$N$B}; we refer to a checkpoint by the number of tokens it has seen.} %
For Pythia we use steps 1k, 4k, 16k, 32k, 64k and 143k \og{(2B, 8B, 32B, 64B, 128B and 286B tokens).} %

\begin{table}[h]
  \centering\small
  \setlength{\tabcolsep}{5pt}
  \begin{tabular}{@{}rrrrr@{}}
    \toprule
    Tokens & 1B & 7B & 13B & 32B \\
    \midrule
    9B    & --   & 2k   & 1k  & 1k  \\
    17B   & --   & --   & 2k  & 2k  \\
    21B   & 10k  & 5k   & --  & --  \\
    42B   & 20k  & 10k  & 5k  & 5k  \\
    84B   & 40k  & 20k  & 10k & 10k \\
    210B  & 100k & 50k  & 25k & 25k \\
    630B  & 300k & 150k & 75k & 75k \\
    1678B & 800k & --   & --  & --  \\
    \bottomrule
  \end{tabular}
  \caption{Training steps of the OLMo-2 stage-1 checkpoints used for fine-tuning, by training tokens (OLMo-2-0425-1B, OLMo-2-1124-7B, OLMo-2-1124-13B, OLMo-2-0325-32B).}
  \label{tab:ckpts}
\end{table}

\paragraph{Fine-tuning.}
\nw{\og{The learning rate is $10^{-4}$ for LoRA (rank 8, $\alpha=16$, all linear layers), $10^{-3}$ for ReFT (a rank-8 additive edit of the two operand states at the constrained layers, base model frozen) and $10^{-5}$ for full fine-tuning.} All three use AdamW without weight decay, batch size 16 and 3 epochs.} %
\og{At 42B tokens the helix prior raises accuracy from 18\% to 28\% with LoRA, from 14\% to 29\% with ReFT and from 17\% to 39\% with full fine-tuning ($\lambda=20$); the shuffled helix gives 7--11\%.} %
\og{The self-timed prior of \S\ref{sec:window}, with running average $\bar a_t=0.95\,\bar a_{t-1}+0.05\,\mathrm{acc}_t$, was run on addition at 42B, 84B and 210B tokens and is on for 95--100\% of the steps at 42B and 10--11\% at 210B.} %
Learning-curve runs use 7{,}000 training pairs for 10 or 30 epochs and measure test accuracy every 50 or 100 steps. %
\og{In the 10-epoch runs of \S\ref{sec:window} at the earliest checkpoint, a digit code also ends at 99.9\% but learns about 1.5 times more slowly than the helix.} %
\og{All arms of a seed share the data order and the initialisation of the trainable parameters, so comparisons are paired.}
\og{In fine-tuning, the prior acts on the 90 prompts ``\texttt{Q:~$x$}'' at every step; with causal attention, the state of $x$ there equals that of the first operand in ``\texttt{Q:~$x$~+~$b$~=~}''.} %
\og{At the 21B-token checkpoint, the model with the prior reaches 90\% held-out accuracy \og{after 1{,}150 steps (3{,}050 without it), and the model without it then catches up} (98.9\% against 99.8\% with the prior after 10 epochs; \S\ref{sec:window}).} %
Subtraction uses $a\ge b$ and multiplication $ab\le999$, so every answer is a single token. %
\og{With the new operator $a\,\#\,b=(a+b)\bmod100$ of \S\ref{sec:window}, the gain of the prior is $+20.5$ points at 42B tokens and $-6.6$ at 210B.} %
\og{For subtraction, the helix raises accuracy from 14.4\% to 19.0\%, from 27.3\% to 39.7\% and from 50.2\% to 57.2\% at 21B, 42B and 84B tokens, and gives 97.5\% and 99.6\% against 96.4\% and 100.0\% at 210B and 630B (3 seeds).} %

\paragraph{Pythia.}
The GPT-NeoX tokenizer \citep{black2022gptneox} attaches the leading space to numbers, so the prompt is ``\texttt{Q:~$a$~+~$b$~=}'' and the answer is the single token \nw{``\texttt{~$a{+}b$}''. The numbers up to 198 are single tokens.} %
Data, splits, targets and hyper-parameters are those of OLMo-2.
\og{We exclude Pythia-2.8B: on the Hugging Face Hub, its weight files at steps 1k, 4k and 16k have the same hash, as do those at 64k and 143k, and the five checkpoints give identical results.} %

\paragraph{Concept domains.}
Weekdays and months are trained on ``\texttt{Q:~$\langle$item$\rangle$ plus/}\allowbreak\texttt{minus $k$ days/}\allowbreak\texttt{months is}'' with $k\in\{1,\dots,40\}$, letters on ``\texttt{Q:~F plus 3 letters is}'', and number words and Roman numerals (I--XVI) on ``\texttt{Q:~seven + twelve =~}''.
\og{Every item is a single token in the same position of every prompt, so its state in ``\texttt{Q:~$\langle$item$\rangle$}'' equals its state in every task prompt.}
All combinations are shuffled once and \nw{split 60/40 into training and test. Runs use 280 optimizer steps}, batch size 16, learning rate $10^{-4}$ and $\lambda=20$. %
\og{Because the split is over (item, operation, $k$) and $k$ runs up to 40, every weekday and month test problem has a training problem with the same item and the same shift modulo 7 or 12.} %
\og{The circle still helps months in a rerun with a strict split (40\% of the (item, shift mod $n$) classes held out, 3 seeds): 59.4\% vs.\ 28.7\% without a prior at 210B tokens. About 70\% of these test problems have their reverse \og{(answer month, inverse operation, same $k$)} in training; the month gain of Table~\ref{tab:helped}, \og{with a new strict split per seed (Appendix~\ref{app:screen}), holds also where the reverse was never trained}. For weekdays the strict test is at floor (all arms near or below chance), so it cannot tell whether the circle helps.} %

\paragraph{Transfer formats.}
In the multi-format experiments the training batches rotate over ``\texttt{Q:~$a$~+~$b$~=~}'', ``\texttt{Compute $a$ + $b$. Result:~}'', ``\texttt{The sum of $a$ and $b$ is~}'' and ``\texttt{$a$ plus $b$ equals~}'' (and the corresponding subtraction formats). %
\og{\og{With this rotation, the gain carries over to the held-out format below} ($+4.3$ to $+7.0$ points for addition and subtraction at 42B and 84B tokens; \S\ref{sec:window}).} %
The held-out GPT-3 format is ``\texttt{Q:~What is $a$ plus $b$?\textbackslash n\textbackslash nA:~}'', restricted to items of the GPT-3 arithmetic set \citep{brown2020gpt3} with operands 10--99 \og{whose pair does not occur in training.} %

\paragraph{Compute.}
\og{The experiments, including failed, cancelled and preempted jobs, used at least 3{,}350 GPU hours: about 2{,}940 for pretraining from scratch, 210 for fine-tuning and representation analysis and 190 for the toy networks.} %

\FloatBarrier   %
\section{Target Geometries and Losses}
\label{app:targets}
\begin{table}[t]
  \centering\footnotesize
  \setlength{\tabcolsep}{3pt}
  \begin{tabular}{@{}llll@{}}
    \toprule
    Concept & Target & Setting & Effect of the prior \\
    \midrule
    Numbers    & helix       & PT 53M             & 18.4 $\rightarrow$ \nw{69.9\%$^{a}$} \\
               &             & FT 7B, 9B          & 90\% $5\times$ \nw{sooner$^{b}$} \\
    Months     & circle      & FT 7B, 210B        & $+21.6$ points$^{c}$ \\
    Weekdays   & \nw{circle$^{d}$} & FT 7B, 210B       & $+32.6$ points \\
 \sx{Dimensions} & \sx{lattice} & \sx{PT 53M, 6 priors$^{e}$} & \sx{$+33.9$ points} \\ %
 \sx{Cities} & \sx{sphere} & \sx{PT 53M, 6 priors$^{e}$} & \sx{$+6.3$ points} \\ %
 \sx{Countries} & \sx{sphere} & \sx{PT 53M, 6 priors$^{e}$} & \sx{$+5.8$ points} \\ %
 \sx{$\mathbb{Z}_6^2$ group} & \sx{torus} & \sx{PT 53M, 6 priors$^{e}$} & \sx{$+22.0$ points} \\ %
 \sx{Grid sums} & \sx{grid} & \sx{PT 53M, 6 priors$^{e}$} & \sx{$+19.7$ points} \\ %
 \nw{Vector sums} & \nw{coordinates} & \nw{toy, 6.7\% of table} & \nw{0/5 $\rightarrow$ 5/5 learn}$^{f}$ \\ %
    \bottomrule
  \end{tabular}
 \caption{\sx{Concepts a manifold prior helped (all 36 fine-tuning domains: Table~\ref{tab:domains36}). PT: pretraining from scratch; FT: fine-tuning OLMo-2-7B at the given checkpoint. Effects are in held-out accuracy, in PT averaged over training. $^{a}$Means over 8 seeds, each arm at its own best learning rate. $^{e}$Six priors of Table~\ref{tab:comb}; the targets encode the answers. $^{f}$Toy network whose target encodes the whole table: seeds reaching 95\% (\S\ref{sec:limits}). Notes $b$--$d$ and details: Appendix~\ref{app:capnotes}.}}   %
  \label{tab:helped}
\end{table}

\begin{table}[h]
  \centering\small
  \setlength{\tabcolsep}{3pt}
  \begin{tabular}{@{}l>{\raggedright\arraybackslash}p{0.42\linewidth}>{\raggedright\arraybackslash}p{0.25\linewidth}@{}}
    \toprule
    Target & Coordinates of $x$ & Tests \\
    \midrule
    \geo{helix} & Eq.~\eqref{eq:helix}: $\cos,\sin(2\pi x/T)$, $T\in\{2,5,10,100\}$; $x/100$ & the full manifold \\
    \addlinespace[2pt]
    \geo{digit} & one-hot(tens) $\oplus$ one-hot(units) & a place-value code \\
    \addlinespace[2pt]
    \geo{sep\_ce} & no target shape; leave-one-out class-mean cross-entropy of $x \bmod 2,5,10$ and of the tens digit & separability only \\
    \addlinespace[2pt]
    \geo{helix\_nonsep} & \geo{helix} plus fixed noise in the $T{=}2,5,10$ coordinates, with no power at any helix frequency & Fourier spikes without separability \\
    \addlinespace[2pt]
    \geo{helix\_coarse} & the $T{=}100$ pair and $x/100$ only & neither (control) \\
    \addlinespace[2pt]
    \geo{$\ast$\_shuf} & the same coordinates on a fixed permutation of 10--99 & loss strength only \\
    \bottomrule
  \end{tabular}
 \caption{Target geometries for the numbers 10--99. \og{\geo{helix} has 8 dimensions.} \geo{sep\_ce} replaces the CKA term and uses $\lambda=5$. %
  }
  \label{tab:geoms}
\end{table}

\paragraph{Number targets.}
\og{\og{For \geo{helix\_nonsep}, the fixed Gaussian noise on the five $T{=}2,5,10$ coordinates of \geo{helix} is made orthogonal to the constant vector and to all helix columns and scaled to a standard deviation of 3 per column.}} %
\og{The CKA of \geo{helix\_nonsep} to the helix is 0.51.} %

\paragraph{\og{Noisy helices.}}
\geo{helix\_n}$k$ adds fixed Gaussian noise with standard deviation $k/100$ to the standardised helix coordinates of every number \nw{($k=25,50,100,200$). The CKA of these targets with the clean helix is 0.95, 0.83, 0.57 and 0.29, respectively.} %
\og{\og{This noise lowers the gain of the prior strictly monotonically}, from $+11.8$ to $-6.9$ points at 42B tokens and \fz{from $+71.4$ to $-3.5$ at 9B tokens with 7{,}000 pairs (a separate set of runs, 3 seeds)} (\S\ref{sec:geometry}).} %

\paragraph{Separability loss.}
\og{\og{\geo{sep\_ce} computes its leave-one-out nearest-class-mean cross-entropy on the same centred states as the CKA term, with squared distances to the class means divided by $\tau=0.1\cdot\frac{1}{n}\sum_i\lVert h_i\rVert^2$, and averages it over its four labellings and over layers.}} %

\paragraph{Concept targets.}
\og{Number words and Roman numerals use \geo{helix} and \geo{digit} of their values.}
\og{The answer-circle target of \S\ref{sec:geometry} assigns the final-token state of a training prompt with answer index $j=(i\pm k)\bmod n$ the point $\bigl[\cos\frac{2\pi j}{n},\sin\frac{2\pi j}{n}\bigr]$, at the layers at $3/4$ and $7/8$ of the depth, on 96 training prompts resampled at every step.}
Its controls place the answers at permuted positions on the circle, $j\mapsto\sigma(j)$, or use equidistant clusters, $e_j$. %

\paragraph{\og{Why a shape loss.}}
\label{app:loss}
\og{Our first loss, $\lVert (h-\mu)R - F_G(x)\rVert^2$ with a ridge read-out $R$ fitted on the base model's mean states of the 90 numbers, \og{did not constrain the shape: 90 points in 4096 dimensions map linearly onto any target, even a shuffled one,} and the targets did not differ in effect (ReFT at 42B tokens: 13.8--15.7\% held-out accuracy, 13.8\% for \geo{none}).} %

\FloatBarrier   %
\section{Further Results on the Benefit Window}
\label{app:tables}
\begin{figure*}[t]\centering\includegraphics[width=\textwidth]{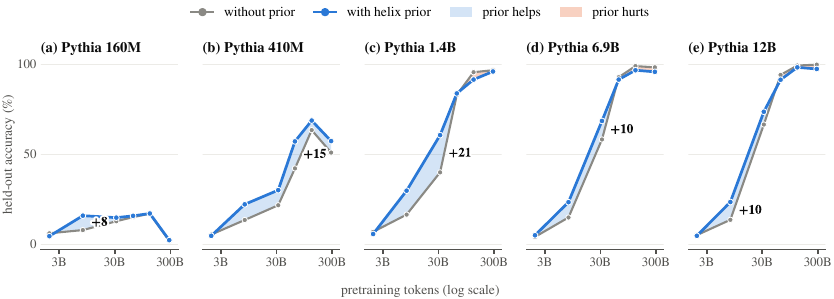}\caption{The benefit window for Pythia 160M--12B on two-digit addition, drawn as in Figure~\ref{fig:window}a--d (3 seeds).}\label{fig:pythia}\end{figure*} %
\paragraph{\og{Amount of training data.}} %
\og{The gain of the helix on OLMo-2-7B is $+4.2$, $+30.2$ and $+70.4$ points at 9B tokens with 1{,}500, 4{,}000 and 7{,}000 training pairs (3 epochs), $+4.5$, $+48.1$ and $+70.4$ at 21B, $+4.3$ and $+12.5$ at 42B with 400 (11 epochs) and 1{,}500 pairs, and $+0.8$, $+4.1$ and $+10.6$ at 84B with 150 (30 epochs), 400 and 1{,}500 pairs (5 seeds with 1{,}500 pairs, otherwise 3; paired $p\le0.014$ except for the two smaller sets at 84B).} %
\sx{A first stage that raises the separability of $x \bmod T$ does not increase the gain of a helix applied after it, so the lower edge of the window is not set by the maturity of the number representation.}   %
\og{With 7{,}000 pairs at 9B tokens, \geo{none} reaches 11.3\% and the helix 81.7\%; a shuffled helix gives 7.6\%, the digit code 36.7\%, \og{and \geo{none} with learning rate $3\times10^{-4}$, 12.6\%}.} %
\paragraph{\og{Three-digit sums at maturity.}} %
\og{On three-digit sums \og{($a,b\in[100,499]$), never trained on,} the helix lowers accuracy at 630B tokens from 68.1\% to 57.7\% for OLMo-2-7B (shuffled helix: 54.6\%) and from 90.4\% to 80.7\% for OLMo-2-32B, the same as with the shuffled helix.} %
\paragraph{\og{General ability.}} %
\og{After fine-tuning OLMo-2-7B with or without the prior in four settings (9B--84B tokens), each of HellaSwag, ARC-Easy, ARC-Challenge and LAMBADA (500 items each, standard error about 2 points) stays within 0.7 points of the base model, and the language-modelling loss on held-out FineWeb-Edu text within 0.011.} %

\FloatBarrier   %
\section{Further Results: Fine-Tuning, Toy Grokking and the Atlas}
\label{app:mult}
\paragraph{\og{Multiplication.}} %
\og{We tested targets in which multiplication is closer to a translation (\geo{log}: $\log x$; \geo{factor}: $\log x$, the exponents of 2, 3, 5 and 7, and a one-hot units digit; its shuffled twin; \geo{helix\_factor}: \geo{helix} $\oplus$ \geo{factor}) and per-place codes (\geo{digit}, the digit-wise Fourier basis of \citet{bai2025multiplication}, the two circles of FoNE \citep{zhou2025fone}, and the helix with that basis), on OLMo-2-7B at 42B and 84B tokens (3 seeds).} %
\og{At $\lambda=20$ every target lowers accuracy in the training format, by 6.5 to 20.3 points, the shuffled codes most, and the harm shrinks with the loss weight until, at $\lambda=1$, there is neither harm nor gain.} %
\og{On held-out GPT-3-format pairs, \geo{factor} and \geo{helix\_factor} gain up to $+6.2$ and $+8.5$ points at 84B tokens ($p\le0.026$).} %
\og{FoNE does improve multiplication, but it builds a Fourier code into the input of models trained mostly from scratch, whereas our prior reshapes the representation of a pretrained model.} %
\paragraph{\og{Targets from a mature model.}} %
\label{app:mature} %
\og{\og{The target \geo{main} is the Gram matrix of the final OLMo-2-7B model's own states of the same numbers at the same layers.}} %
\og{At 9B, 42B and 84B tokens, \geo{main} gives 4.4\%, 17.6\% and 59.4\% held-out accuracy, against 5.3\%, 18.0\% and 56.6\% for \geo{none} and 9.6\%, 28.3\% and 66.4\% for the helix, although it reaches a CKA of 0.99 with the final model, whose own geometry has a CKA of only about 0.76 with the helix.} %
\og{Constraining only the input-embedding rows of the number tokens gives 6.9\%, 25.6\% and 69.3\% at 9B, 42B and 84B tokens \fz{(\geo{none} in these runs: 4.6\%, 17.7\%, 58.5\%)}, and the CKA with the helix at layer 16 rises from 0.69--0.74 to 0.78--0.81.} %
\paragraph{\og{Three-digit sums, held-out operands and inference.}} %
\label{app:further} %
\gn{We fine-tune the 9B-token checkpoint on addition of numbers from 0 to 999, with 7{,}000 training pairs that cover 0.7\% of the table. The model without a prior memorises:} \nw{it is correct on every training pair and on 0.8\% of the held-out pairs}\og{.} %
With the helix prior, the pre-registered area under the held-out accuracy curve rises by 13.0 points \nw{(3 of 3 seeds)}\og{, and final accuracy by 28.6 points, almost all of it on sums of 1000 or more, whose units digit is a separate token (51.0\% against 0.6\%), not on sums below 1000} (6.3\% against 0.9\%). %
\og{The rule for the units digit generalises to unseen pairs, while the higher digits are about as accurate as answering with the sum of the nearest training pair \fz{(a look-up that gets 6.1\% of either kind of sum exactly right;} for the 42B checkpoint, see below).} %
\og{In fine-tuning on three-digit sums at the 42B checkpoint, the model with the prior reaches 45.8\% against 11.0\% without it, and 28.6\% on sums below 1000, which is above the accuracy of the nearest-pair look-up.} %
\og{When 15 operand values never occur in fine-tuning, the prior raises accuracy on problems that contain them from 16.2\% to 29.3\% at 42B tokens and from 48.4\% to 60.1\% at 84B, and to 24.6\% and 57.1\% when the prior never sees these values.} %
\og{Snapping the operand states, fitted as $h\approx\mu+F_G(x)W_G$ (helix or digit code), to the ideal point of $x$ at one layer never improves accuracy in OLMo-2 1B, 7B and 13B (at most $+2$ points) and lowers it more for more capable models; the ideal point of a wrong number collapses accuracy wherever the model can add (7B at 210B tokens: from 49\% to 2\%), and a norm-matched random direction has no effect.} %

\paragraph{\og{Resampling the geometry batch.}}
\label{app:resample}
\og{For the answer circle of \S\ref{sec:geometry}, resampling its batch of 96 training prompts at every step \fz{makes it help weekdays and months on test problems with equivalent trained shifts ($\lambda=5$, 5 seeds, $p=0.001$ each)}; with one fixed batch, the model memorises the batch, and the circle lowers test accuracy at 210B tokens (weekdays $-37.5$, months $-32.0$ points).} %
\og{The gain comes from the circle's geometry: with the answers at permuted positions on the circle the target hurts ($-30.4$ and $-20.4$ points, $p\le0.016$), and equidistant clusters do not help significantly ($+0.9$ and $+11.6$, $p\ge0.13$).} %
\paragraph{\og{Toy grokking.}} %
\label{app:toy} %
\og{In the modular-addition setting of \citet{nanda2023grokking} ($p=113$, 30\% of the pairs for training, a one-layer transformer, AdamW with learning rate $10^{-3}$ and weight decay 1, 25k full-batch steps, 5 seeds), a prior $1-\cka(W_E[0{:}p],F)$ pulls the embedding of the residues towards a circle of frequency 1 or a target with frequencies 1--5, for 2000 steps from a chosen onset or throughout.} %
\og{With frequencies 1--5, the earlier the prior is applied the faster the model groks, up to about four times faster than without it, as in \citet{howe2026windows} and \citet{kazanskii2026tame}, although with the prior on throughout or from the first step one of five seeds does not grok; the single circle helps only after the training set is memorised and, applied from the first step, slows memorisation so that most seeds fail to grok.} %
\paragraph{\og{The manifold atlas.}} %
\label{app:atlas} %
\og{\og{We measured how well the final-token states of short prompts} (``\texttt{The month of March}'') linearly decode target coordinates for 36 concept domains, across the pretraining of OLMo-2-7B (11 checkpoints), OLMo-2-1B (9) and Pythia-6.9B (5), and at the final checkpoints of eight models (these three and OLMo-3-7B, SmolLM2-1.7B, Qwen2.5-7B, Qwen3-8B-Base and Mistral-7B; \citealp{olmo3,allal2025smollm2,qwen25,qwen3,jiang2023mistral}\og{); ability was measured with 3-shot ``$k$ steps after'' tests.}} %
\og{\textbf{Representations before abilities.} For numbers, number words, ordinals, months, hours and units, half of the final level is reached between 4B and 21B tokens by the geometry, between 84B and 210B by the stepping ability for numbers and number words, and between 630B and 3.9T for ordinals, months and hours (bootstrap probability that ability comes later $\ge0.95$).} %
\og{\textbf{Decodable coordinates, not one shape.} Excluding two domains whose geometry comes from the tokenizer (years, musical pitch), 27 of 34 domains are decodable ($R^2$ above a permutation null by $\ge0.3$, $p<0.05$) in all eight models (19--30 under other thresholds or templates); but wrong-shape targets are often almost as decodable.} %
\paragraph{\og{Multiple priors at once.}} %
\label{app:rainbow} %
\og{We fine-tune OLMo-2-7B on addition, weekday and month arithmetic together, with a helix for numbers and a circle for each calendar domain, alone or combined (Table~\ref{tab:rainbow}; LoRA, $\lambda=20$ per prior or $\lambda=20/K$ for $K$ priors).} %
\begin{table*}[t] \centering\small \setlength{\tabcolsep}{3pt} \begin{tabular}{@{}lrrrrrr@{}} \toprule & \multicolumn{3}{c}{84B tokens} & \multicolumn{3}{c}{210B tokens} \\ \cmidrule(lr){2-4}\cmidrule(l){5-7} Prior & add & week & month & add & week & month \\ \midrule none (accuracy)        & 48.6 & 38.6 & 32.5 & 96.5 & 48.6 & 55.2 \\ helix                  & \up{+11.3} & $-1.2$ & $-4.6$ & \dn{-3.8} & $-8.6$ & $-8.9$ \\ weekday circle         & $+0.3$ & \up{+15.5} & $-0.9$ & $+0.2$ & $-1.4$ & $-9.4$ \\ month circle           & $-2.0$ & $-0.6$ & $+9.7$ & $-0.1$ & $-10.3$ & \up{+20.8} \\ week + month           & $-1.0$ & $+9.9$ & $+18.0$ & $+0.0$ & $-0.4$ & \up{+18.0} \\ all three              & \up{+9.7} & \up{+10.3} & $+7.6$ & \dn{-6.7} & $+6.0$ & $+18.5$ \\ all three, $\lambda/3$ & \up{+12.5} & \up{+13.6} & \up{+12.3} & \dn{-3.7} & \up{+12.1} & \up{+22.0} \\ all three, shuffled    & $-11.4$ & $-21.9$ & $-15.5$ & $-14.6$ & $-33.6$ & $-30.2$ \\ \bottomrule \end{tabular} \caption{Multiple priors at once (OLMo-2-7B, gain over \geo{none} in points; addition on problems whose two orders were never trained; 5 seeds for the key arms). Shaded: $p<0.05$, paired $t$-test. Weekday and month tests measure transfer across equivalent shifts (Appendix~\ref{app:details}). } \label{tab:rainbow} \end{table*} %
\paragraph{\og{Operation-specific priors.}} %
\label{app:opcond} %
\og{\og{With addition, subtraction and multiplication trained together, the prior acts on one shared representation (the separate ``\texttt{Q:~$x$}'' prompts) or only on the second operand of one operation's problems.} Logarithmic or factor-based priors within multiplication problems do not help multiplication (Table~\ref{tab:opcond}), and their harm on multiplication problems whose other order was trained is matched by their shuffled versions (not in the table), so it is a cost of the constraint rather than of the shape.} %
\begin{table*}[t] \centering\small \setlength{\tabcolsep}{4pt} \begin{tabular}{@{}lrrrrrrrr@{}} \toprule & \multicolumn{4}{c}{42B tokens} & \multicolumn{4}{c}{84B tokens} \\ \cmidrule(lr){2-5}\cmidrule(l){6-9} Prior & add & sub & mult strict & mult all & add & sub & mult strict & mult all \\ \midrule none (accuracy)        & 13.5 & 26.2 & 14.8 & 49.9 & 41.6 & 53.6 & 29.2 & 68.4 \\ shared helix           & \up{+7.3} & \up{+9.7} & \dn{-2.3} & \dn{-9.2} & \up{+13.2} & \up{+6.6} & \dn{-2.4} & \dn{-3.0} \\ helix in add./sub.     & \up{+18.8} & \up{+11.7} & \dn{-2.2} & $-1.5$ & \up{+19.3} & \up{+11.1} & $-2.0$ & $-1.0$ \\ shuffled, in add./sub. & $-3.3$ & $-6.6$ & $0$ & $-0.5$ & $-7.4$ & \dn{-16.5} & $-1.0$ & $-1.9$ \\ log in mult.           & $+2.0$ & \up{+2.7} & $+1.0$ & \dn{-7.9} & $-1.5$ & $-0.2$ & $+1.5$ & \dn{-5.8} \\ factor in mult.        & \up{+4.4} & \up{+5.1} & $-0.5$ & \dn{-10.8} & $+2.1$ & $+1.6$ & \dn{-4.1} & \dn{-11.2} \\ \bottomrule \end{tabular} \caption{Operation-specific priors (OLMo-2-7B, gain over \geo{none} in points; 5 seeds for the key arms, 3 for the shuffled arm). ``strict'': problems whose two orders were never trained. Shaded: $p<0.05$ (blue: gain, orange: loss), paired $t$-test over seeds, not corrected for multiple comparisons. } \label{tab:opcond} \end{table*} %

\FloatBarrier   %
\section{Pretraining from Scratch: Details and Further Results}
\label{app:pretraining}

\fz{The setup, evaluation and held-out problems below apply to all pretraining runs. From ``Site of the prior'' on, the results use earlier runs, with operands 10--99 unless stated otherwise and learning rate $10^{-3}$ for all arms, at which the model without a prior is not tuned; Appendix~\ref{app:pt999} gives the main runs with operands 0--999.} %

\paragraph{Setup.}
The models follow the OLMo-2 architecture with 6 layers, width 384, 6 heads, feed-forward width 1536 and tied input and output embeddings (53M parameters, of which 38.5M are embeddings). %
\og{The larger models have 10 layers of width 512 (93M parameters), \sx{11 of width 640 (136M),} 12 of width 768 with 12 heads and feed-forward width 3072 (190M), 19 of width 832 (294M), \sx{20 of width 960 (391M),} 24 of width 1024 with 16 heads and feed-forward width 4096 (505M; 403M without the embeddings)\sx{, 29 of width 1152 (731M)} and 36 of width 1280 with 20 heads and feed-forward width 5120 (1.07B; 944M without the embeddings). \og{Apart from the learning rate (and the number of tokens in Appendix~\ref{app:pttpp}), they share the setup below.}} %
\fz{Unless stated otherwise, all models are trained on 300M tokens (sequences of 512 tokens, batch 32) with AdamW \citep{loshchilov2019adamw} (learning rate $10^{-3}$ unless it is chosen per arm,} $\beta=(0.9,0.95)$, weight decay 0.1), 1\% linear warm-up, cosine decay to 10\% and gradient clipping at 1.0. %
Each sequence of FineWeb-Edu text \nw{is followed by either four addition problems (7\% of the tokens) or one (2\%), in one of four formats} \og{(those of the multi-format fine-tuning runs, Appendix~\ref{app:details})}. %
The prior uses $\lambda=5$ \nw{at layers 2 and 4. The self-timed prior is on while} an exponential moving average (decay 0.98) of the training-batch accuracy on the first problem of each sequence is below 0.5. %

\paragraph{Evaluation in training position.}
In training, every problem follows about 500 tokens of text.
\og{When held-out problems are given as bare prompts at the start of a sequence, a position never seen in training, accuracy is low and erratic, even when the model solves them almost perfectly in training position.} %
\og{All pretraining results in this paper therefore evaluate held-out problems after a held-out chunk of text, so that they end where training problems end.} %

\paragraph{Held-out problems.}
\fz{The runs with operands 10--99 hold out both orders of about 500 unordered pairs (985 problems, on which held-out accuracy is measured) and also hold out 300 problems whose other order is trained.} %
\og{In the main runs with operands 0--999, each seed draws its own held-out split, initialisation and data order, so the four held-back seeds evaluate the chosen learning rates in independent runs.} %
\og{\og{The held-out problems of the held-back seeds overlap those of the selection seeds in about a quarter of the curve-average weight (a sixth at 1.07B), mostly sums of two one-digit operands or of a one- and a two-digit operand. On sums of a two- and a three-digit operand or of two three-digit operands, which almost never overlap, the operand prior still leads on all four held-back seeds in every setting of \S\ref{sec:faster} where the held-back seeds confirm the gain.}} %

\paragraph{\og{Site of the prior.}} %
\og{\og{The earlier site comparisons of \S\ref{sec:place} use 7\% arithmetic, 4 seeds and priors applied throughout; without a prior, the model reaches 90\% after a median of 201M tokens, and a shuffled helix on the operands lowers the curve average from 63.9\% to 56.9\%.}} %
\paragraph{\og{Tokenization.}} %
\og{\og{With 7\% arithmetic (4 seeds), splitting numbers into single-digit tokens (all other tokens unchanged) lets the model without a prior reach 90\% after a median of 31M tokens instead of 205M; with priors on ``\texttt{Q:~$x$}'' prompts, whole-number tokens need 37--41M (windowed priors) and digit tokens 20--33M (every prior). With whole-number tokens and operands up to 999, priors covering the whole range (two helix variants and one circle per digit) raise the curve average from 38.2\% to 84--88\%, and no arm reliably adds four-digit numbers.}} %
\paragraph{\og{Targets read out of a trained model.}} %
\og{The target can also be the number geometry of the trained OLMo-2-7B} (layer 16, its first 8 principal components or all dimensions), applied to the operands in context. %

\gn{\og{Three targets with a CKA of about 0.62 to the helix (the 8-component read-out, a noisy helix and a units-plus-tens one-hot code) \fz{raise the curve average by 24--26 points over no prior (the noisy helix and the one-hot code on the same hardware)} and are indistinguishable from each other;}} the clean helix is 5--7 points higher. %
The one-hot code spans 99\% of the variance of the standardised helix, so this comparison tests whether the helix frequencies need extra weight, not whether they are needed at all. %

\FloatBarrier   %
\section{Pretraining with Operands up to 999}
\label{app:pt999}

\paragraph{Setup.}
\og{\og{The main pretraining runs use the setup of Appendix~\ref{app:pretraining} with operands 0--999 and the \geo{helix3} target (18{,}310 steps, 75 evaluations), all on one GPU type.}} %
\nw{\og{Held-out problems are scored by exact match of the multi-token answer, and the curve average weights the six combinations of operand lengths equally.}} %
\og{A fixed pool of 525 pairs (5\% of the arithmetic lines) recurs in training.} %
\nw{\og{Tokens to 90\% are medians over seeds, counting a seed that never reaches 90\% at the budget, and refer to the first crossing unless stated otherwise.}} %
\og{Paired tests use seeds that share initialisation, data order and split; GPU nondeterminism makes their trajectories diverge before the prior is switched on, so pairing removes only the variance from these sources. With 4 seeds, an exact sign-flip test cannot go below $p=0.125$, so ``4 of 4'' rests on the paired $t$-test.} %

\paragraph{Main results.}
Table~\ref{tab:pt999} lists the main arms.
\og{\fz{With 7\% arithmetic, a shuffled helix on the operands, matched in steps}, gradient norm and language-modelling cost lowers the curve average by 23.6 points (0 of 8 seeds), while the correct one raises it by 32.4 (8 of 8).} %

\begin{table*}[t]
  \centering\small
  \setlength{\tabcolsep}{4pt}
  \begin{tabular}{@{}lrrrrr@{}}
    \toprule
    Arm & Curve avg. & Final & Final 3-digit & Tok.\ to 90\% & $\Delta$bits/byte ($\times10^{-3}$) \\
    \midrule
    \multicolumn{6}{@{}l}{\emph{2\% arithmetic (main setting), 8 seeds}} \\
    none                                    & 9.4  & 17.2 & 2.8  & -- (0/8) & -- \\
    operands in context, from step 0$^\dagger$ & 59.1 & 85.2 & 53.3 & -- (1/8) & $+45.8$ \\
    web text, from step 0$^\dagger$            & 47.5 & 75.6 & 32.7 & -- (0/8) & $+46.6$ \\
    web text, window $[0.05,0.4)$            & 65.5 & 95.5 & 79.2 & 228 (7/8) & $+10.7$ \\
    \midrule
    \multicolumn{6}{@{}l}{\emph{7\% arithmetic (mechanism setting), 8 seeds}} \\
    none                                    & 56.2 & 93.9 & 76.6 & 200 (7/8) & -- \\
    operands in context, throughout         & 88.6 & 99.4 & 96.5 & 57  & $+17.6$ \\
    operands in context, self-timed         & 87.9 & 99.4 & 97.5 & 57  & $+16.6$ \\
    web text, window $[0.05,0.4)$            & 85.8 & 100.0 & 99.8 & 80  & $+7.2$ \\
    web text, self-timed                    & 80.2 & 98.4 & 91.5 & 126 & $+56.6$ \\
    operands in context, shuffled helix     & 32.6 & 65.5 & n/r & -- (1/8) & $+16.2$ \\
    \bottomrule
  \end{tabular}
 \caption{Pretraining with operands 0--999 (53M parameters, learning rate $10^{-3}$, $\lambda=5$): \og{curve-average and final held-out accuracy, final accuracy on three-digit operands (\%)}, median tokens (M) to 90\% (in parentheses: seeds that reach it), and change in bits per byte on the full validation set (2M tokens) relative to no prior. $^\dagger$Self-timed priors that, with 2\% arithmetic, stay on for 68--100\% (operands) and 100\% (web text) of the steps. \og{Final accuracies hold for this learning rate only.} n/r: not reported.}
  \label{tab:pt999}
\end{table*}

\paragraph{Shuffled control with 2\% arithmetic.}
\og{\og{In the window, the correct helix beats the shuffled one by $+63.7$ points on the operands and $+57.0$ in web text (8 of 8 seeds).}} %
\og{On the operands, the shuffled and the correct prior are on for the same steps and are matched in fit, gradient and language-modelling cost. In web text only the steps match, because the shuffled target is fitted more closely but pulls less.} %
\og{With four times as many tokens, the shuffled helix lowers the curve average by 41.4 points, while the correct helix raises it by 18.1.} %

\paragraph{Each arm at its own learning rate.}
\og{For the main setting we chose each arm's learning rate by curve-average accuracy on seeds 0--3 and repeated the chosen pair on seeds 4--7 (Table~\ref{tab:pt2main}).} %
\nw{Compared with the model without a prior at its best learning rate for language modelling, the prior costs} 0.0065--0.0079 bits per byte at 53M and 0.0097 at 190M after 300M tokens, and no cost is detectable after 1.2B tokens. %

\begin{table*}[t]
  \centering\small
  \setlength{\tabcolsep}{5pt}
  \begin{tabular}{@{}lrrrrr@{}}
    \toprule
    Setting & No prior: lr, curve avg. & Prior: lr, curve avg. & $\Delta$ seeds 0--3 [95\% CI] & $\Delta$ seeds 4--7 & $\Delta$ 8 seeds \\
    \midrule
    53M, 300M tokens  & $8\times10^{-3}$, 23.1   & $2\times10^{-3}$, 69.8 & $+46.7$ [$+37.0$, $+56.5$] & $+56.2$ & $+51.5$ (8/8) \\
    190M, 300M tokens & $4.8\times10^{-3}$, 26.8 & $6\times10^{-4}$, 78.6 & $+51.7$ [$+37.1$, $+66.4$] & $+47.1$ & $+49.4$ (8/8) \\
    \nw{505M, 300M tokens} & \nw{$4.8\times10^{-3}$, 24.7} & \nw{$6\times10^{-4}$, 76.7} & \nw{$+52.0$ [$+42.1$, $+62.0$]} & \nw{$+53.9$} & \nw{$+53.0$ (8/8)} \\
    53M, 1.2B tokens  & $2\times10^{-3}$, 69.2   & $10^{-3}$, 85.1        & $+15.9$ [$+11.9$, $+19.9$] & $+17.1$ & $+16.5$ (8/8) \\
    \bottomrule
  \end{tabular}
 \caption{The main setting (2\% arithmetic, operands 0--999) with each arm at its own best learning rate: \og{curve-average held-out accuracy (\%) without a prior and with the windowed prior on the operands (seeds 0--3)}, and paired differences on the selection seeds, the held-back seeds and all 8 seeds (4 of 4 positive in each half). \nw{At 505M (total parameters), the learning rates are the best of the grids we tried (near-ties broken by bits per byte); the grid without the prior has no lower rate, and the prior's curve average is highest at its lowest rate. \og{Cluster, GPU and training code differ across sizes, so the rows are not a size comparison (\S\ref{sec:faster})}.}} %
  \label{tab:pt2main}
\end{table*}

\paragraph{Each arm at its own learning rate, 7\% arithmetic.}
\gn{\og{Table~\ref{tab:pt7lr} repeats the procedure above with 7\% arithmetic at two model sizes.}} %
\nw{\og{On all 8 seeds, the prior first reaches 90\% 3.65 and 3.87 times sooner at 53M and 190M.}} %
\og{\og{At 505M, over learning rates from} $3\times10^{-4}$ to $1.92\times10^{-2}$ (seeds 0--3), the pre-registered rule chooses $9.6\times10^{-3}$ \og{without a prior, although it is only 4.6 points above} $4.8\times10^{-3}$ (3 of 4 seeds, not significant).} %
\og{\og{Against it, the prior at $1.2\times10^{-3}$ leads by 27.4 points on the selection seeds (4 of 4), and the four held-back seeds agree (4 of 4).}} %
\og{\og{At 1.07B, the runs use B200 GPUs instead of L40S.}} %
\og{\og{On seeds 0 and 1 only, the pre-registered rules choose} $2.4\times10^{-3}$ without a prior (of $1.2\times10^{-3}$ to $9.6\times10^{-3}$) and $6\times10^{-4}$ with it (of $3\times10^{-4}$ to $1.2\times10^{-3}$; a near-tie broken by bits per byte); Table~\ref{tab:pttpp} gives the 505M and 1.07B results at 300M tokens.} %

\begin{table*}[t]
  \centering\small
  \setlength{\tabcolsep}{3pt}
  \begin{tabular}{@{}lrrrrrrr@{}}
    \toprule
    Size & No prior: lr, avg. & Prior: lr, avg. & $\Delta$ seeds 0--3 [95\% CI] & $\Delta$ 8 seeds & 50\% & 90\% (first / stays) & stays 95\% \\
    \midrule
    53M  & $8\times10^{-3}$, 65.8   & $2\times10^{-3}$, 89.1   & $+23.4$ [$+16.9$, $+29.8$] & $+22.8$ (8/8) & 4.75$\times$ & 3.13 / 3.11$\times$ & 1.53$\times$ \\
    190M & $4.8\times10^{-3}$, 64.3 & $1.2\times10^{-3}$, 89.8 & $+25.4$ [$+12.6$, $+38.2$] & $+23.5$ (8/8) & 4.56$\times$ & 4.05 / 4.40$\times$ & 2.89$\times$ \\
    \bottomrule
  \end{tabular}
  \caption{7\% arithmetic, operands 0--999, each arm at its own best learning rate (only the learning rate was tuned): curve-average held-out accuracy (\%, seeds 0--3), paired differences, and how many times sooner the prior reaches 50\% and 90\% (first crossing; ``stays'': from when accuracy stays above the threshold; seeds 0--3). Held-back seeds 4--7: $+22.2$ and $+21.6$, 4 of 4 each.}
  \label{tab:pt7lr}
\end{table*}

\paragraph{\og{Weight, site and start.}} %
\og{With $\lambda=1$ (7\% arithmetic, self-timed priors from step 0, seeds 0--3), \fz{the curve average is 89.2\% in context and 86.7\% in web text, so the gap between the two sites shrinks to 2.5 points (not significant).} \og{At $\lambda=5$, the ratio of prior to language-modelling gradients when the window opens is 3.9 in web text, 1.9 in context and 1.5 on prompts.} In Figure~\ref{fig:pretraining}b, start and duration are confounded, because all windows end at 40\% of training.} %
\paragraph{\og{Window or self-timed prior.}} %
\og{Self-timed priors that also start at 5\% of training and switch off when the training accuracy on the weakest operand-length combination reaches 0.5 (or 0.9) are compared with the window in four settings (the default, learning rate $5\times10^{-4}$, layers 1--6 and 600M tokens; all with 7\% arithmetic, prior in context, seeds 0--3). \og{The pre-registered rule chooses the window (worst-case regret 0, against 1.96 and 1.19 points of curve-average three-digit accuracy)}; the threshold-0.5 prior is 1.6--2.0 points lower in each setting (14 of 16 seed pairs), within the registered margin of 2 points, \og{and their bits per byte differ by} $-0.0008$ (90\% CI $-0.0030$ to $+0.0014$).} %
\paragraph{\og{General ability.}} %
\og{We evaluate the final models zero-shot on the full BLiMP, LAMBADA, HellaSwag, ARC-Easy and PIQA sets and compare each prior with the model without a prior at the same learning rate and seeds (Table~\ref{tab:downstream}); \og{the comparisons and a one-point margin were registered} before any score was seen.} %
\nw{BLiMP is scored on whole sentences without a leading space, with ties counted as wrong, \nw{\gn{\og{so our BLiMP scores are not comparable with other reports.}} For LAMBADA we report greedy accuracy on all target tokens, for HellaSwag normalised accuracy, and for ARC-Easy and PIQA plain accuracy.}} %
\og{At the same learning rate, the validation loss with the prior is higher in 4 of 4 seeds at 505M with $3\times10^{-4}$ (by 0.012--0.020 nats per token) and not detectably different at 505M with $1.2\times10^{-3}$.} %

\begin{table*}[t]
  \nwcolor   %
  \centering\small
  \setlength{\tabcolsep}{3pt}
  \resizebox{\textwidth}{!}{%
  \begin{tabular}{@{}lccccc@{}}
    \toprule
    Comparison (same learning rate, four paired seeds) & BLiMP & LAMBADA & HellaSwag & ARC-Easy & PIQA \\   %
    \midrule
    Number prior, 505M, $1.2\times10^{-3}$ & $-0.72$ $[-2.39,+0.94]$ U & $-0.63$ $[-0.92,-0.34]$ N$^{*}$ & $-0.11$ $[-0.90,+0.67]$ N & $-0.46$ $[-2.48,+1.56]$ U & $+0.46$ $[-0.32,+1.24]$ N \\   %
    Number prior, 505M, $3\times10^{-4}$   & $+0.30$ $[-0.55,+1.15]$ N & $-0.20$ $[-0.80,+0.40]$ N & $-0.10$ $[-0.52,+0.31]$ N & $+0.01$ $[-1.51,+1.53]$ U & $+0.08$ $[-1.20,+1.36]$ U \\   %
    \midrule
    Six priors, 190M, $6\times10^{-4}$  & $+0.14$ $[-0.61,+0.89]$ N & $-0.27$ $[-0.76,+0.21]$ N & $-0.24$ $[-0.72,+0.24]$ N & $+0.08$ $[-0.80,+0.97]$ N & $-0.27$ $[-0.95,+0.40]$ N \\   %
    Six priors, 505M, $6\times10^{-4}$  & $+0.04$ $[-0.56,+0.63]$ N & $-0.04$ $[-1.13,+1.05]$ U & $-0.16$ $[-0.83,+0.50]$ N & $+0.07$ $[-1.43,+1.58]$ U & $+0.04$ $[-0.33,+0.41]$ N \\   %
    Six priors, 1.07B, $6\times10^{-4}$ & $-0.31$ $[-1.88,+1.27]$ U & $-0.24$ $[-1.30,+0.82]$ U & $+0.22$ $[-0.12,+0.56]$ N & $+0.56$ $[-2.25,+3.36]$ U & $-0.45$ $[-2.01,+1.11]$ U \\   %
    \bottomrule
  \end{tabular}}
 \caption{\fz{Zero-shot benchmarks after pretraining with priors in the window:} the number prior at 505M (7\% arithmetic, seeds 0--3) and the six priors at 190M--1.07B (seeds 16--19), minus no prior at the same learning rate and seeds; mean with 95\% $t$-interval (points). N: non-inferior with a margin of one point; U: undetermined with 4 seeds (not a difference below one point). $^{*}$A small cost: the interval lies below zero. No correction for multiple comparisons; only final checkpoints are evaluated.} %
  \label{tab:downstream}
\end{table*}

\FloatBarrier   %
\section{\nw{Model Size or Amount of Data}}
\label{app:pttpp}
\begin{figure*}[t]
  \centering
  \includegraphics[width=\textwidth]{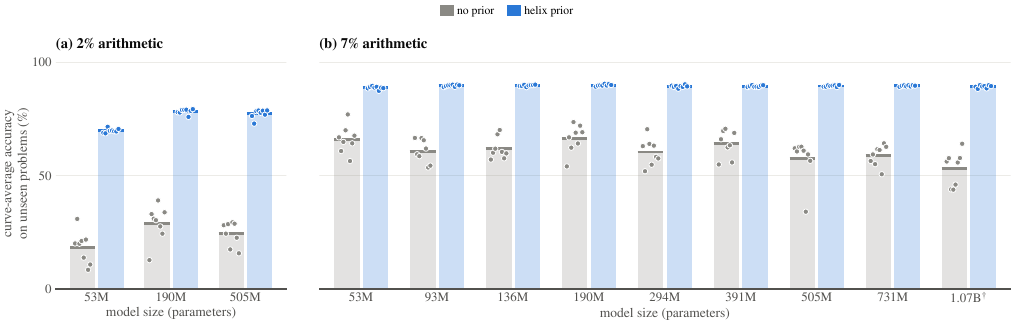}
 \caption{\og{\fz{The gain at several model sizes (dots: seeds; bars: means). (a,~b)~Curve-average held-out accuracy, each arm at its own best learning rate, 8 seeds;} $^{\dagger}$on B200 GPUs, with the learning rates chosen on 2 seeds. Details: Appendix~\ref{app:capnotes}.}} %
  \label{fig:scaling}
\end{figure*}
\og{Each size is trained at a fixed ratio of tokens to non-embedding parameters: 0.318 and 0.745 with 7\% arithmetic, 0.745 with 2\%. The sizes are 93M, 190M, 294M, 505M and 1.07B (Appendix~\ref{app:pretraining}), on L40S GPUs except 1.07B (B200) and the 2\% runs (A100).} %
\og{Each arm's learning rate is chosen on seeds 0--3 (0--1 at 1.07B) \fz{by the rules of Tables~\ref{tab:pt2main} and~\ref{tab:pt7lr}}; four held-back seeds then run at these rates.} %
\og{The primary metric, the difference in curve-average accuracy, was registered before any verdict was read; \og{the log ratio of times to 50\% must agree in sign}.} %

\begin{table}[t]
  \nwcolor   %
  \centering\footnotesize
  \setlength{\tabcolsep}{2.5pt}
  \setlength{\tabcolsep}{2pt}   %
  \begin{tabular}{@{}llrrrrr@{}}
    \toprule
    Size & tpp & Tokens & No prior & Prior & Gain (8 seeds) & Held-back \\
    \midrule
    \multicolumn{7}{@{}l}{\emph{7\% arithmetic}} \\
    93M   & 0.318 & 13.3M  & 1.5  & 0.9  & $-0.6$ (0/8) & floor \\
    93M   & 0.745 & 31.3M  & 3.8  & 12.5 & $+8.8$ (6/8) & floor \\
    190M  & 0.318 & 36.0M  & 4.7  & 37.8 & $+33.1$ (8/8) & $+26.9$ \\
    190M  & 0.745 & 84.4M  & 9.5  & 73.9 & $+64.4$ (8/8) & $+65.7$ \\
    294M  & 0.318 & 66.9M  & 7.9  & 66.4 & $+58.5$ (8/8) & $+56.3$ \\
    294M  & 0.745 & 156.8M & 37.9 & 84.0 & $+46.2$ (8/8) & $+49.1$ \\
    505M  & 0.318 & 128.0M & 19.8 & 82.8 & $+63.0$ (8/8) & $+59.8$ \\
    505M  & 0.745 & 300M   & 57.5 & 89.7 & $+32.2$ (8/8) & $+36.9$ \\
    1.07B & 0.318 & 300M   & 53.3 & 89.4 & $+36.1$ (8/8) & $+33.5$ \\
    1.07B & 0.745 & 703.1M & 79.3 & 92.3 & $+13.0$ (8/8) & $+12.6$ \\
    53M   & 21.2  & 300M   & 66.1 & 88.9 & $+22.8$ (8/8) & $+22.2$ \\
    190M  & 2.65  & 300M   & 66.5 & 90.0 & $+23.5$ (8/8) & $+21.6$ \\
    \midrule
    \multicolumn{7}{@{}l}{\emph{2\% arithmetic}} \\
    190M  & 0.745 & 84.4M  & 4.8  & 34.9 & $+30.1$ (8/8) & $+30.7$ \\
    294M  & 0.745 & 156.8M & 7.9  & 62.7 & $+54.8$ (8/8) & $+55.1$ \\
    505M  & 0.745 & 300M   & 24.4 & 77.4 & $+53.0$ (8/8) & $+53.9$ \\
    \bottomrule
  \end{tabular}
 \caption{Curve-average held-out accuracy (\%) at a fixed ratio of tokens to non-embedding parameters (tpp), each arm at its own learning rate, \og{8 seeds; gain: mean paired difference \og{(seeds higher with the prior)}; held-back: seeds 4--7. ``Floor'': excluded from the trend test by a pre-registered rule.} }
  \label{tab:pttpp}
\end{table}

\nw{Table~\ref{tab:pttpp} lists every cell.} %
\og{With 2\% arithmetic, the rise of the gain with size along the 0.745 line is not fully shown, \fz{because at 190M and 294M the median model without a prior never reaches 50\%, so the log ratio of times to 50\% cannot be computed.}} %
\og{With 7\% arithmetic, within each size, the gain first rises with the number of tokens (in cells of up to about 128M tokens) and then falls, with a peak at roughly 70M to 130M. Each arm needs roughly the same number of tokens to reach 50\% at every size where it does (about 90M to 120M without a prior, 17M to 24M with it), so the 0.745 cells lie on the falling side.} %
\sx{At 300M tokens, two pre-registered extensions added 93M, 136M, 294M, 391M and 731M (294M also on B200 GPUs), each arm with 8 seeds at its own learning rate chosen on seeds 0--3 (Table~\ref{tab:pt300}). The second registered a test over the eight L40S sizes: an ordinary least-squares slope of the 8-seed mean gains on the log of non-embedding parameters, shown if (a)~its seed-bootstrap 90\% interval excludes zero, (b)~the two halves of the seeds agree in sign and (c)~the 90\% interval from the residual spread of the sizes excludes zero. All three hold (\S\ref{sec:faster}; halves $+1.16$ and $+2.17$). Four of the eight sizes (53M, 93M, 190M and 505M) had been seen before registration, and (c) holds only just and only with 136M ($[-0.11,+3.60]$ without it): leaving out any one size keeps the slope positive ($+1.18$ to $+2.08$), but (c) then holds in 3 of the 8 cases. The registered secondary test, on the three never-seen sizes alone, has a power of 0.47--0.63, so its failure to exclude zero does not show that these sizes lack the increase. None of 18 swaps to a learning rate within the tie band changes the sign of the slope. At 294M the two GPU types give the same gain (L40S minus B200: $-0.12$ points, 90\% interval $[-5.2,+4.7]$). Over all ten cells, with an indicator for the GPU type, the slope is $+1.94$ (descriptive). With the prior, the median seed reaches 50\% 4.2 to 5.2 times sooner at each size.} %
\begin{table}[t]   %
  \ifdraftmarks\color{revbrown}\fi
  \centering\footnotesize
  \setlength{\tabcolsep}{2pt}
  \begin{tabular}{@{}lrrrrrr@{}}
    \toprule
    Size & tpp & lr ($10^{-3}$) & No prior & Prior & Gain & Held-back \\
    \midrule
    53M           & 21.2  & 8 / 2       & 66.1 & 88.9 & $+22.8$ & $+22.2$ \\
    93M           & 7.15  & 2.4 / 1.2   & 60.9 & 89.8 & $+28.9$ & $+31.0$ \\
    136M          & 4.16  & 4.8 / 0.6   & 62.0 & 89.9 & $+27.8$ & $+27.9$ \\
    190M          & 2.65  & 4.8 / 1.2   & 66.5 & 90.0 & $+23.5$ & $+21.6$ \\
    294M$^{\dagger}$& 1.43  & 9.6 / 0.6   & 60.8 & 89.9 & $+29.1$ & $+30.2$ \\
    294M          & 1.43  & 9.6 / 0.6   & 60.5 & 89.5 & $+29.0$ & $+31.1$ \\
    391M          & 1.02  & 9.6 / 0.6   & 64.1 & 89.6 & $+25.5$ & $+26.9$ \\
    505M          & 0.745 & 9.6 / 1.2   & 57.5 & 89.7 & $+32.2$ & $+36.9$ \\
    731M          & 0.487 & 9.6 / 0.6   & 59.0 & 89.7 & $+30.7$ & $+30.0$ \\
    1.07B$^{\dagger}$& 0.318 & 2.4 / 0.6   & 53.3 & 89.4 & $+36.1$ & $+33.5$ \\
    \bottomrule
  \end{tabular}
  \caption{\sx{Curve-average held-out accuracy (\%) at 300M tokens with 7\% arithmetic. tpp: tokens per non-embedding parameter; lr: without / with the prior, chosen on seeds 0--3 (0--1 at 1.07B); gain: mean paired difference, with 8 of 8 seeds higher with the prior in every row; held-back: seeds 4--7. $^{\dagger}$On B200 GPUs; the other rows ran on L40S GPUs and form the pre-registered test.}}
  \label{tab:pt300}
\end{table}
\sx{One seed without a prior at 505M learns late (curve average 34.2, against a median of 60.9 in its cell); leaving out single seeds gives slopes from $+1.32$ to $+1.90$.} %
\paragraph{\sx{With 2\% arithmetic.}}   %
\sx{With 2\% arithmetic and 300M tokens (A100 GPUs; a setting we do not compare with 7\%), the helix prior raises held-out accuracy averaged over training by 48.0, 51.6, 49.9, 52.3 and 53.0 points at 53M, 93M, 190M, 294M and 505M parameters (8 of 8 seeds at each size). The gain increases with log size, by 1.27 points per $e$-fold of non-embedding parameters (pre-registered test: seed-bootstrap 90\% interval $[0.02,2.49]$, both halves of the seeds positive, residual interval $[0.13,2.41]$, two-sided $p=0.08$); on the four seeds not used to choose learning rates, the slope is 1.41. This is a partial confirmation: we had seen three of the five sizes before registering the test, and given them a gain that does not grow would still read as an increase 16--38\% of the time (likelihood ratio about 1.7--3.8); with the 294M learning rate one step away, the verdict flips but the slope stays between 1.04 and 1.54; at a fixed number of tokens, size moves together with tokens per parameter (21.2 to 0.745); and the rise above 190M comes from the model without the prior getting weaker (28.0\%, 26.0\% and 24.4\% at 190M, 294M and 505M) while the model with the prior stays at about 78\%.} %

\FloatBarrier   %
\section{A Screen of Priors over 36 Domains}
\label{app:screen}
We fine-tuned OLMo-2-7B at 42B and 210B tokens on the stepping or comparison task of each of the 36 atlas domains, with the domain's target as prior, its shuffled twin, or neither (LoRA, $\lambda=20$, 3 seeds with seed-specific splits). %
\og{In all 216 runs the prior moved the item representations towards its target (CKA at least 0.91), so absent effects are not failures to apply it.} %
\og{Ten cells gain robustly (Holm-corrected; \citealp{holm1979}), but eight are anchor comparisons whose targets imply at least 99\% of the held-out answers (the training problems imply at most 12.5\%) and whose gains fall almost entirely on problems training cannot imply: they read only as facts injected by the target or as faster recall of known facts.} %
\og{Among stepping tasks, whose target gives only the order of the items, the only robust gains are the two known windows, replicated with a new split per seed: numbers at 42B (partly from faster fitting of the training problems) and months at 210B.} %
The four robustly harmful cells are equally harmed by the shuffled target, so the harm is a cost of the constraint rather than of the particular geometry. %
\og{Table~\ref{tab:domains36} lists all 36 domains.}

\begin{table*}[p]
  \centering\footnotesize
  \setlength{\tabcolsep}{6pt}
  \begin{tabular}{@{}llllrrr@{}}
    \toprule
    Domain & Prompt & Target & Task & Dec. & 42B & 210B \\
    \midrule
    \multicolumn{7}{@{}l}{\emph{Numbers}} \\
    Numbers & The number $x$ & helix & step & 8/8 & \up{+41.6} & $-1.1^{c}$ \\
    Number words & The number $x$ & helix & step & 8/8 & $-6.9$ & \dn{-18.5}$^{c}$ \\
    Ordinals & It came $x$ & helix & step & 8/8 & $+2.0$ & $-10.1$ \\
    Years$^\ddagger$ & In the year $x$ & helix & step & 8/8 & \dn{-14.0} & \dn{-2.4}$^{c}$ \\
    \multicolumn{7}{@{}l}{\emph{Cycles}} \\
    Weekdays & The day was $x$ & circle & step & 8/8 & $0.0^{f}$ & $-1.0^{f}$ \\
    Months & The month of $x$ & circle & step & 8/8 & $+1.2^{f}$ & \up{+21.6} \\
    Hours & The time was $x$ & circle & step & 8/8 & $-2.0$ & $+4.1$ \\
    Zodiac signs & The zodiac sign $x$ & circle & step & 6/8 & $+1.2^{f}$ & $+1.2^{f}$ \\
    Chinese zodiac & The Year of the $x$ & circle & step & 3/8 & $+0.6^{f}$ & $+0.6^{f}$ \\
    Musical keys & A piece in the key of $x$ & chromatic & step & 2/8 & $+0.6^{f}$ & $-0.3^{f}$ \\
    Pitch$^\ddagger$ & The note $x$ & chromatic & step & 2/8 & $+0.3^{f}$ & $-0.6^{f}$ \\
    Compass & The wind blew from the $x$ & circle & anchor & 8/8 & $+3.2$ & $-0.6^{c}$ \\
    Colors & The color $x$ & hue circle & anchor & 8/8 & \up{+4.0} & $+0.8^{c}$ \\
    \multicolumn{7}{@{}l}{\emph{Magnitudes and scales}} \\
    Animal weight & The weight of a $x$ & log line & order & 8/8 & $-0.1^{c}$ & $-0.1^{c}$ \\
    Country population & The population of $x$ & log line & order & 8/8 & $0.0^{c}$ & $0.0^{c}$ \\
    Kinship & My $x$ & generation & order & 8/8 & $0.0^{c}$ & $+0.2^{c}$ \\
    Scalar adjectives & It was $x$ & within scale & order & 3/8 & $-2.3$ & $+3.8$ \\
    Units & One $x$ & within scale & order & 8/8 & $+0.5^{c}$ & $-2.5^{c}$ \\
    \multicolumn{7}{@{}l}{\emph{Orders}} \\
    Letters & The letter $x$ & line & step & 8/8 & \dn{-13.8} & $-10.2$ \\
    Greek letters & The Greek letter $x$ & line & step & 8/8 & $0.0^{f}$ & $-1.0^{f}$ \\
    Planets & The planet $x$ & line & step & 8/8 & $0.0^{f}$ & $+15.1^{f}$ \\
    Chemical elements & The element $x$ & line & step & 7/8 & $0.0^{f}$ & $0.0^{f}$ \\
    Geologic periods & The geologic period $x$ & line & step & 8/8 & $0.0^{f}$ & $-8.8^{f}$ \\
    US presidents & President $x$ & line & order & 8/8 & $+0.5^{c}$ & $-0.2^{c}$ \\
    Historical figures & The historical figure $x$ & rank & order & 8/8 & $-0.7^{c}$ & $+0.3^{c}$ \\
    Military ranks & The military rank of $x$ & line & order & 8/8 & $+6.2$ & $0.0^{c}$ \\
    Life stages & The life stage $x$ & line & order & 6/8 & $0.0^{c}$ & $0.0^{c}$ \\
    Metric prefixes & The metric prefix $x$ & line & order & 8/8 & $0.0$ & $+1.6$ \\
    \multicolumn{7}{@{}l}{\emph{Places}} \\
    Countries & The country of $x$ & sphere & anchor & 8/8 & \up{+2.4}$^{c}$ & $+1.5^{c}$ \\
    Cities & The city of $x$ & sphere & anchor & 8/8 & \up{+3.7} & $+1.7$ \\
    US cities & The American city of $x$ & plane & anchor & 8/8 & $+0.2^{c}$ & $-0.1^{c}$ \\
    Landmarks & The famous landmark $x$ & sphere & anchor & 8/8 & \up{+11.9} & \up{+14.1} \\
    Stars & In the night sky, the star $x$ & sphere & anchor & 1/8 & \up{+18.7}$^{f}$ & \up{+42.2}$^{f}$ \\
    Emotions & I feel $x$ & plane & order & 8/8 & $+0.1^{c}$ & $+0.5^{c}$ \\
    \multicolumn{7}{@{}l}{\emph{Taxonomies}} \\
    Animals & The animal $x$ & tree & anchor & 8/8 & $+0.8^{c}$ & $+0.8^{c}$ \\
    Languages & The language $x$ & tree & anchor & 8/8 & \up{+4.3}$^{c}$ & $-0.9^{c}$ \\
    \bottomrule
  \end{tabular}
 \caption{\og{The 36 domains of the atlas (Appendix~\ref{app:atlas}) and of the screen. Prompt: the template whose final-token state represents item $x$. \og{Dec.: final models (of eight) in which} the target coordinates are linearly decodable. Task: \emph{step}, ``$k$ steps after $x$'' (the target carries no answer); \emph{order}, which of two items is larger or earlier; \emph{anchor}, which of two items is closer to a third (the target coordinates imply 93.6--100\% of the held-out answers). 42B / 210B: \og{gain in held-out accuracy (points; OLMo-2-7B)}.} Shaded: robust gain or loss (Holm-corrected problem- and split-level tests, 3 of 3 seeds, stable when leaving out one seed). $^{f}$/$^{c}$: accuracy without the prior at floor / ceiling. $^\ddagger$Geometry largely determined by the tokenizer.}
  \label{tab:domains36}
\end{table*}

\section{Can a Prior Make an Unlearnable Task Learnable?}
\label{app:walls}
\begin{table*}[t]
  \nwcolor   %
  \centering\small
  \setlength{\tabcolsep}{4pt}
  \begin{tabular}{@{}>{\raggedright\arraybackslash}p{3.0cm}>{\raggedright\arraybackslash}p{4.6cm}p{9.0cm}@{}}
    \toprule
    Cost & Setting & Effect \\
    \midrule
    \multicolumn{3}{@{}l}{\textit{Of the prior itself}} \\[1pt]
    \textbf{General ability} & PT 505M, \sx{the number prior} (window, two learning rates); PT 190M--1.07B, the six priors; same learning rate and seeds without the prior, 4 seeds & On five zero-shot benchmarks, \textbf{no significant decline larger than 1 point} in \sx{10} comparisons; \sx{in 6 the interval excludes a loss above 1 point and 4 are undetermined; one small cost: LAMBADA,} $-0.63$ $[-0.92,-0.34]$. Six priors, same rate, 190M--1.07B: \textbf{no benchmark changes by more than 0.6 points}; 0 worse, 9 non-inferior, 6 undetermined of 15$^{a}$ \\   %
    \addlinespace[3pt]
    \textbf{Language modelling} & PT, window from 5\%, each arm at its best learning rate & \gn{\textbf{0.6--0.9\% higher bits per byte}} after 300M tokens, not detectable after 1.2B; at the same learning rate, validation loss higher in 4 of 4 seeds in one of the two comparisons above \\   %
    \addlinespace[6pt]
    \multicolumn{3}{@{}l}{\textit{At the wrong time}} \\[1pt]
    \textbf{Prior from the first step} & PT 53M, 7\%; PT 505M, 7\% & \textbf{About eight times} the language-modelling cost of the window (0.064 against 0.008 bits per byte); at 505M a self-timed prior on web text from the first step ($\lambda=5$, learning rate not tuned for it) is 1.1--3.3 points lower on all five benchmarks$^{b}$ \\
    \addlinespace[3pt]
 \textbf{Model that has learned the task} & FT OLMo-2, addition and seven domains & \textbf{Gain negative once capability exceeds about 80\%} (all four OLMo-2 sizes); $-1.3$ points above 90\% across seven \nw{domains; \nw{on three-digit sums (not in the fine-tuning data), about 10 points lower at 630B tokens}} \\
    \addlinespace[6pt]
    \multicolumn{3}{@{}l}{\textit{With the wrong geometry}} \\[1pt]
 \textbf{Wrong geometry} & PT 53M & Shuffled helix, at the same learning rate as the model without a prior: \textbf{\boldmath$-23.6$ points} of curve-average accuracy (7\%, same \nw{weight, lower in 8 of 8 seeds),} $-41.4$ with four times the tokens \nw{(2\%, lower in 4 of 4 seeds)} \\ %
    \addlinespace[3pt]
 \textbf{Geometry the computation does not use} & FT OLMo-2-7B, three operations & One shared helix lowers multiplication by \textbf{11--12 points} \nw{at 42B and 84B tokens and still by 8.8 at 210B} (in a separate experiment, $-1.0$ to $-1.5$ when applied only to the second operand of addition and subtraction problems, against $-3.0$ to $-9.2$ for a shared helix) \\ %
    \addlinespace[6pt]
    \multicolumn{3}{@{}l}{\textit{With several priors}} \\[1pt]
 \textbf{Multiple priors} & \sx{PT 53M--1.07B, the six priors of Table~\ref{tab:comb}} & \sx{Switched off after 40\% of training, \textbf{the torus and the grid end up to 3.0 and 1.6 points below} the best models without priors at 53M; kept on until the end, all six end above them (post hoc). At the same learning rate, \textbf{0.3\% higher bits per byte} at 190M and \textbf{1.4\%} at 505M (4 of 4 seeds each), 0.0\% on average at 1.07B} \\   %
    \bottomrule
  \end{tabular}
 \caption{\nw{What a manifold prior costs. PT: pretraining from scratch; FT: fine-tuning. $^{a}$BLiMP, LAMBADA, HellaSwag, ARC-Easy and PIQA; all intervals are in Table~\ref{tab:downstream}, details in Appendix~\ref{app:capnotes}. $^{b}$This configuration also has the largest language-modelling cost we measured.}}   %
  \label{tab:costs}
\end{table*}
\og{Criteria, fixed in advance, for a task that a prior makes learnable: without a prior, a learning-rate and weight-decay sweep with the best setting inside the range, and four times the budget; a learning-rate sweep for the prior; a target that places no answer, though it may give a structure from which the table follows; failure of a shuffled or random target; held-back seeds; one GPU type for all arms. For the held-out numbers below, the weight-decay criterion was added after a first read-out.} %

\paragraph{Numbers never seen in arithmetic.}
\og{In 53M pretraining with 7\% arithmetic (L40S GPUs, learning rate $2\times10^{-3}$), we first held out 50 three-digit numbers per seed from every arithmetic line. Without a prior, models still answer 21\% of the problems with one held-out operand; a prior that reaches the held-out numbers lifts this to 94\% (on the embedding rows) while it is on, but within about 2k steps after the window closes the coordinates it supplied are overwritten and accuracy falls.} %

\og{\fz{In a second batch (A100 GPUs, 54 held-out numbers per seed, six in each hundred), the prior stays on} from 5\% of training to the end, and the read-out is the final accuracy on problems whose held-out operand lies between 700 and 999 (seeds 0--3 to choose, 4--7 to confirm).} %
\og{The prior on the embedding rows reaches 98.9\%, against 1.7--5.0\% without a prior in four settings and 3.8\% with a prior on the operands (against the model without a prior at its two main learning rates and against the operand prior, $+95$ to $+97$ points; 4 of 4 held-back seeds, $p<10^{-5}$, Holm-corrected).} %
\og{Given instead the coordinates of a seen number, a held-out number's row nearly coincides with that number's (cosine 0.97), and in 7 of 8 seeds 94--99\% of its problems are answered as if it were that number (44\% in the remaining seed).} %
\og{Settings without a prior added after the first read-out (learning rate $10^{-3}$, weight decay 0.3, and four times the tokens at $2\times10^{-3}$) reach at most 11.3\%; with four times the tokens, the model without a prior exceeds the pre-registered floor of 10\% in 2 of 4 seeds (11\%), so this case does not meet the criteria either.} %

\paragraph{Few training examples.}
\og{In a one-layer toy transformer, tuned training without a prior stays at the floor for modular addition at 5\% and 10\% of the table and for S4 at 20\% and 30\% (S4 cells reclassified after the runs); priors that fix only the manifold's shape stay there too, as do their controls, so we cannot tell whether they help.} %
\og{A prior that also fixes which item goes where raises the area under the held-out curve on the two modular-addition walls (torus target; 0.156 and 0.255, against 0.014 and 0.018 without a prior; 5 of 5 seeds, $p=0.004$), but neither it nor a tuned matrix prior for S4 with up to four times the budget (0 of 5 seeds) makes the network learn these walls; the torus prior was not tested with a shuffled target, its own learning-rate range, the full budget or held-back seeds.} %

\paragraph{\nw{A wall that a prior lifts.}}
\sx{Beyond Section~\ref{sec:limits}: without a prior, in every setting that trains, the network memorises the training problems and stays near chance on the others (at most 3.5\%, chance 1.1\%); with the prior, the seed that reaches 95\% only intermittently ends at 90.1\%, so 4 of 5 end at or above 95\%, and every run with the prior has a larger area under the held-out accuracy curve than every run of each comparison (pre-registered one-sided Mann--Whitney test, $p=0.004$). Without a prior, the model learns the same table from 22\% of it. The screen covers five tasks, three training fractions and two kinds of target.}   %
\og{\og{For the vector task of \S\ref{sec:limits}}, the model is a one-layer transformer (width 128, 4 heads, MLP width 512, no layer normalisation) trained full-batch in fp32 with AdamW ($\beta=(0.9,0.98)$) at a constant learning rate; the 5 seeds were registered before the first screen, and all runs use L40S GPUs.} %
\nw{Without a prior we ran 28 settings, each with 5 seeds and four times the standard budget (200k steps): learning rates $3\times10^{-3}$, $10^{-2}$ and $3\times10^{-2}$ at weight decay 0, 0.01 and 0.03; $10^{-3}$ to $10^{-1}$ at 0.1, 1 and 3; and $10^{-3}$ to $3\times10^{-2}$ at 10, so the grid is not a full product.} %
\og{At weight decay 0 the training loss becomes exactly zero in fp32 and the weights stop changing, and at $3\times10^{-2}$ with weight decay up to 0.03 training diverges or oscillates.} %
\og{The prior pulls the input embeddings of the 125 vectors towards their exponent coordinates.} %
\og{The pre-registered Mann--Whitney test ($p=0.004$) treats the arms as independent although they share seeds; as 5 pairs, 5 of 5 gives $p=0.031$.} %

\paragraph{Where training succeeds without a prior.}
\og{In the toy, a network without a prior learns S4 from 60\% of the table in 5 of 5 seeds, and a prior towards a faithful $3\times3$ matrix representation of the group (\citealp{chughtai2023toy}) makes it learn a median of 4.2 times sooner per seed on the same GPU type.} %
\og{Sums and differences of dimension vectors in $\{-1,0,1\}^3$, trained on 30\% of the relations, are learned without a prior in 2 of 5 seeds and with a prior towards the exponent vectors in 5 of 5, more than 27.8 times sooner in the median than the best setting tested. Both targets encode the whole operation.} %
\og{In these cases the priors make the model learn earlier or in more seeds.} %

\FloatBarrier   %
\section{\cy{Domain Priors and the Six-Prior Combination}}   %
\label{app:domains}

\paragraph{\cy{The screen.}}
\sx{We screened priors for 24 concept domains in 53M pretraining (300M tokens, the 2\% setting, number prior off), from months, colors and cities to kinship terms and geological periods. Each sequence carries two problems of the domain, and the prior acts on the last token of each item name in the window $[0.05,0.4)$ with $\lambda=1$. A pre-registered two-stage rule (seeds 0--3, then 4--7; learning rates chosen per arm; Holm correction over the advanced domains) passes countries, cities, people and months, by 3.8, 3.5, 1.5 and 3.9 points of curve-average held-out accuracy (8 of 8 seeds each; the four share their comparison runs). On people both arms stay below a training-frequency shortcut (95.4\%), and on months the gain lies in the window and both arms end below chance. The other 20 domains do not pass; for 13 of the 17 that did not reach the second stage, the model without a prior preferred a learning rate above the prior's grid, so this is no evidence against their priors.}   %

\paragraph{\cy{Replication with a shuffled target.}}
\sx{On fresh seeds 8--15, never used for any choice, each passing domain is trained with its true target, with a shuffled target that permutes only the assignment of coordinates to items, and without a prior. A comparison passes when the 90\% paired bootstrap interval excludes zero, both halves of the seeds are positive, a one-sided $t$-test is below 0.1 after Holm correction over the four domains and at least 6 of 8 seeds are positive. All four pass (Table~\ref{tab:replicate}). The shuffled target answers only about half of the held-out comparisons correctly and falls below the model without a prior, so part of the contrast between true and shuffled targets is the cost of wrong facts. At the end of training, countries gain 8.7 points on the hard subset (held-out comparisons that neither the answer-frequency nor the shorter-option heuristic answers); for cities the interval includes zero.}   %

\begin{table}[t]
  \centering\small
  \setlength{\tabcolsep}{3pt}
  \resizebox{\columnwidth}{!}{\begin{tabular}{@{}lcccc@{}}   %
    \toprule
    Domain & True $-$ none & True $-$ shuffled & Shuffled $-$ none & End \\
    \midrule
    Countries & $+4.8$ (8/8) & $+14.8$ (8/8) & $-10.0$ $[-12.4,-7.2]$ & $+3.7$ (8/8) \\
    Cities    & $+4.1$ (7/8) & $+16.7$ (8/8) & $-13.3$ $[-16.1,-10.5]$ & $+2.8$ \\
    People    & $+1.2$ (8/8) & $+1.9$ (8/8) & $-0.7$ $[-1.1,-0.4]$ & $+0.7$ (6/8) \\
    Months    & $+4.3$ (8/8) & $+4.3$ (8/8) & $-0.02$ $[-0.18,+0.13]$ & $+1.5$ (7/8) \\
    \bottomrule
  \end{tabular}}
 \caption{\cy{Replication on fresh seeds 8--15 (53M; curve-average held-out accuracy, points; seeds positive). True $-$ none: against the model without a prior at the learning rate chosen in the screen. Shuffled $-$ none: at $2\times10^{-3}$, 90\% interval. End: true $-$ none over the last three evaluations. True $-$ shuffled is at $2\times10^{-3}$; on months all arms are below chance (8.3\%).}}
  \label{tab:replicate}
\end{table}

\paragraph{\sx{Re-screen of near misses.}}
\sx{Six domains that missed the screen narrowly (stars, landmarks, compass points, SI prefixes, military ranks and scalar adjectives) were screened again, with learning rates chosen anew on seeds 0--3 and tests on seeds 8--15 by the replication rule. In the five with a verdict, the true target beats the shuffled one by 5.9 to 15.7 points of curve average (6 to 8 of 8 seeds, Holm $p\le0.043$), but no gain over the model without a prior at its own rate is shown ($-1.2$ to $+3.7$ points, Holm $p\ge0.546$).}   %

\paragraph{\sx{Small algebraic structures.}}
\sx{As candidates for the combination, we screened six small structures in the same setting, each written as ``$a$ \emph{op} $b$ is $c$'' with made-up single-token names: the cyclic group of order 12 (target: a circle), $\mathbb{Z}_6\times\mathbb{Z}_6$ (a torus), $\mathbb{Z}_2^5$ under XOR (the hypercube $\{\pm1\}^5$), the dihedral group of order 12 (a faithful $2\times2$ representation), sums and differences on $\{-2,\dots,2\}^2$ (a grid), and Caesar shifts of the capital letters (a circle; the letters, unlike the made-up names, are frequent tokens). Accuracy is the exact match of the answer on held-out facts. One seed screened the learning rates; the three structures whose prior won ran on two more seeds.}   %

\sx{On all three seeds these priors beat the model without a prior at its best rate ($1.6\times10^{-2}$): the grid by 34.5 points of curve average, the hypercube by 20.9 and the torus by 12.6. The grid and the hypercube stay ahead at the end of training (20.3 points in 3 of 3 seeds, 4.6 in 2 of 3). The torus gains only while its prior is on: it reaches 96.5--99.2\% when the window closes, falls to 80.1--85.9\% afterwards and ends 12.2 points below the model without a prior (0 of 3 seeds). Each target encodes the whole table, there is no shuffled target, and the rates on both sides were chosen on the same three seeds. The cyclic group, the dihedral group and the letters stay near chance in both arms.}   %

\paragraph{\sx{Choosing the torus and the grid.}}   %
\sx{The torus and the grid were picked by a rule registered before any replication run and before the combination ran, though most of the re-screen and the torus's screening seed had been seen. A candidate is eligible when, on at least three paired seeds, neither arm is at the floor or ceiling, the model without a prior is not at the top of its learning-rate range, and the gain is positive on average and in at least two thirds of the seeds with a standardised mean (mean over standard deviation) of at least 0.5; the two eligible priors with the largest standardised mean are taken. No re-screened domain is eligible, and the rule takes the grid (9.09) and the torus (3.50) ahead of the hypercube (3.48).}   %

\paragraph{\sx{The six priors.}}
\sx{The torus and grid problems are those of the structure screen, the city and country problems those of the domain screen, and dimension problems are sums and differences of physical dimensions in $\{-1,0,1\}^3$, written with made-up tokens. For dimensions, the torus and the grid each item is one made-up token, and since input and output embeddings are tied, the token on which the prior acts is also the output row from which it is read as an answer. Each of seeds 16--23 trains, on identical data, the six priors at $10^{-3}$ and $2\times10^{-3}$, each prior alone at $2\times10^{-3}$, and the model without priors at six rates from $2\times10^{-3}$ to $6.4\times10^{-2}$ (the highest added after the pre-registered test).}   %

\sx{A manifold gains when the one-sided paired $t$-test against the model without priors at its best rate for that manifold gives $p<0.05$ and at least 6 of 8 seeds are positive. At $2\times10^{-3}$, interference is not detected if the prior alone gains over the model without priors and the mean of six minus one exceeds minus half of that gain, or if a one-sided test for a loss gives $p\ge0.05$; each prior alone gains over the model without priors at that rate by 49.8, 58.6, 7.3, 6.2, 69.4 and 41.4 points on numbers, dimensions, cities, countries, the torus and the grid. The best rate without priors is $8\times10^{-3}$ for numbers, $4\times10^{-3}$ for cities and countries, $1.6\times10^{-2}$ for the torus and $3.2\times10^{-2}$ for dimensions and the grid, inside the range: at $6.4\times10^{-2}$ every manifold is lower. Against each seed's own best rate the six still gain on all six manifolds (8 of 8 seeds, 7 of 8 on the grid), and against the one rate best for the mean of the six ($1.6\times10^{-2}$) by 22.4 points on average ($[19.6,25.2]$). Most of the gain on numbers comes after the window (31.3 of 46.9 points), about half on dimensions and two fifths on cities and countries, and most of the torus and grid gains inside it.}   %

\paragraph{\sx{Speed, peaks and the lift from joint training.}}   %
\sx{Speed is read off the 8-seed mean curves. The threshold is 90\% of the way from chance to the end-of-training accuracy (mean of the last three evaluations) of the model without priors at its best rate; the six priors at $10^{-3}$ reach it 2.1 to 5.0 times sooner (Table~\ref{tab:comb}; with one evaluation every 250 steps, these ratios are conservative), and they reach the end-of-training accuracy itself after 8 to 30\% of training, in each of the 8 seeds. With their priors on, the torus and the grid peak near the close of the window at 98.2 and 83.3\%, above the peaks of the model without priors at any rate (94.6 and 58.5\%). At the end of training the torus, at 90.4\%, is 3.0 and 1.1 points below the model without priors at $1.6\times10^{-2}$ and $3.2\times10^{-2}$.}   %
\sx{The joint lift on dimensions comes mostly after the window: at $2\times10^{-3}$ the six lead the dimensions prior alone by 4.4 points on average over the evaluations inside the window and by 19.9 after it. On the grid the lift, $+7.4$ points, is positive in 4 of 8 seeds. Without the torus and grid priors (their problems and evaluation stay; the 8 seeds of Table~\ref{tab:comb}), joint training lifts dimensions by only 1.5 points ($[-4.8,+7.8]$): 12.0 points less in curve average ($[+3.8,+20.2]$) and 22.0 less at the end of training ($[+8.6,+35.5]$), both in 7 of 8 seeds (paired $t$-test $p=0.010$ and $0.006$), while numbers, cities and countries change by at most 0.4 points.}   %

\sx{Giving the torus and grid priors shuffled targets of the same strength (same data, same six priors, 91\% of the total prior gradient) removes all of the joint lift on dimensions: the shuffled version is no better than the dimensions prior alone ($-4.5$ points averaged over training, 95\% CI $-14.6$ to $+5.5$, where a dose explanation predicts $+13.5$), while the true targets beat it by 18.0 points averaged over training and 34.5 at the end of training (8 of 8 seeds each). With 1.36 times the prior gradient of the run without the torus and grid priors, the shuffled version is not better than that run either ($-6.0$, $-15.8$ to $+3.7$), so the lift comes from the geometry of the torus and grid targets, not from the extra gradient. The shuffled targets permute the rows of the true ones, which keeps their spectrum and norm, and the model fits them (CKA about 0.99 at the close of the window). The 18.0 points compare with a wrong geometry; the size of the effect of the geometry itself is the 12.0 points above. The effect can be attributed only to the two priors together; the shuffled grid prior gets 73\% of the true dose (the torus 98\%); the shuffled runs ran on mixed GPUs, and the four seeds on the same GPU type as their comparison runs give 13.5 instead of 18.0 (4 of 4); and the test covers 53M at $2\times10^{-3}$ only.}   %

\sx{Training one prior alone also raises the CKA of a structurally similar domain with its own target, although their tokens do not overlap: at the close of the window, the grid and dimensions priors raise each other's by 0.16 and 0.18 and the city and country priors by 0.22 and 0.24 (8 of 8 seeds each), against at most 0.06 between families. Dimensions and the grid are both sums and differences on a small lattice, so similar structure and a similar task cannot be separated here.}   %
\paragraph{\sx{What the prior alone loses after the window.}}   %
\sx{We reran the six and the dimensions prior alone at $2\times10^{-3}$ on seeds 16 and 17, saving the weights at the end of the window (step 7,323), at 12,000 and at the end (18,309). After the window the prior alone loses 21.6 and 18.2 points more than the six on the strict held-out dimension questions, but not its encoding of the inputs: a ridge probe at the first item token decodes the lattice coordinates at layers 2--6 in both models (out-of-fold $R^2$ 0.96--0.99 at the end, difference at most 0.012). At the answer position, layers 5--6 decode the coordinates of the correct answer in all four runs at all three checkpoints ($R^2$ 0.957--0.992), also on the strict questions that the prior alone answers wrongly, beyond what the inputs carry (residual $R^2$ 0.53--0.97). Only the final normalisation and the output layer follow layer 6, so the loss lies in the step from this state to the output token; we did not intervene on the output layer. These are two re-drawn pairs of runs, post hoc.}   %

\paragraph{\sx{Without the torus and grid problems after the window.}}   %
\sx{To test whether practice on the torus and grid problems after the window keeps the lift on dimensions, we reran the six at $2\times10^{-3}$ on seeds 16--23 with these problems removed from the first step after the window and the freed positions filled with web text. At the end of training the six then lead the dimensions prior alone by 19.1 points on the strict held-out dimension questions (7 of 8 seeds, $p=0.011$), against 25.0 with the problems kept: a retained share of 0.77 (seed-bootstrap 95\% interval 0.50--0.96). The torus and the grid fall to 0--3.2\% held-out accuracy within about 400 steps of the removal, which shows only that their answers are no longer produced.}   %

\sx{The removal costs nothing for about 4,700 steps and 5.9 points at the end (7 of 8 seeds, $[+1.6,+10.1]$), and the cost is still growing, so the practice carries about a quarter of the lift at this length (bootstrap interval 4--50\%) and may carry more with longer training. Numbers, cities and countries change by at most 0.5 points.}   %

\sx{Because the removal also put dimension problems at the position of the evaluation questions more often (about 0.75 of them, against 0.50), we reran it with each removed problem replaced in place by web text of the same length: the share is 0.82 (95\% interval 0.65--1.06; 20.4 points, 8 of 8 seeds), the change of position has no measurable effect ($-1.3$ points, $[-7.7,+5.4]$), and by the rule written before the rerun the share of 0.77 stands. That the lift needs the torus and grid priors rests on the runs without them and with shuffled targets above; these runs show that the later problems are not its main source.}   %
\paragraph{\sx{A code shared with the torus.}}   %
\sx{On the re-drawn pairs above, we replaced a $k$-dimensional subspace of the residual stream at the answer position by its mean, at the input of one layer, and measured held-out exact match. Each subspace is spanned by the top principal components of one domain's answer class means; the controls are random subspaces and class-mean components with shuffled labels, and the excess of a drop is the drop minus that predicted from the controls by the removed variance. At the end of training, removing the torus subspace at layer 5 ($k=3$) cuts held-out dimensions accuracy by 45.6 and 43.9 points with the six, against 5.0 and 1.5 with the dimensions prior alone (excess $+0.131$ and $+0.105$, $z$ 4.4 and 2.7); city and country subspaces give an excess of at most $+0.05$. The coupling runs both ways (removing the dimensions subspace cuts torus accuracy by 44 and 45 points, against 7 and 10), though against the torus's own shuffled-label control the excess is only $+0.02$ to $+0.06$; dimensions also depend on the grid answer subspace at layer 6 with the six, not with the prior alone. These are two pairs of runs, post hoc and exploratory: the $z$ values come from five draws of each control, the drop saturates with dose so the linear match is first-order, and only the answer position was edited; the runs keep the torus and grid problems, so it is an inference that this code is why the lift survives their removal.}   %

\paragraph{\sx{Training longer.}}   %
\sx{To test whether more training replaces the joint priors on dimensions, we trained the dimensions prior alone at $2\times10^{-3}$ and the model without priors at $3.2\times10^{-2}$, its best rate for dimensions, for twice and four times as many steps on seeds 16 and 17, with the same window; warm-up and cosine schedule stretch with the run, so after the window the learning rate is higher than at the original length. The prior alone declines with this schedule (seed 16 at step 18,250: 78, 34 and 14\% on the strict held-out dimension questions at the three lengths) and loses structure rather than being undertrained: it stays near 100\% on the dimension questions seen in training and rises to the six priors' level on numbers.}   %

\sx{On all held-out dimension questions, the model without priors matches the six priors already at twice the length (83.9 against 83.0\%); where it matches them on the strict questions, at four times, its CKA with the dimension target is 0.53, against 0.97 (a post hoc read). Seeds 16 and 17 are the prior alone's two best runs at the original length, so their decline includes some regression to the mean; the eight-seed mean at that length (52.3\%) and the CKA are the reference. The longer runs used A100 and H100 GPUs and the original ones B200, a difference far smaller than these effects.}   %
\sx{The separation between the six priors kept on until the end (below) and the model without priors trained four times as long holds on dimensions and the grid within each GPU type (exact one-sided permutation $p=1/45=0.022$ for each). It is a post hoc comparison across two batches, with two seeds and one rate for the longer arm. On the torus, cities and countries, the model without priors does as well or better at $1.6\times10^{-2}$ at the original length, so there the margins over each seed's best rate without priors (4.1, 2.5 and 2.1 points, 8 of 8 seeds) are the comparison.}   %

\paragraph{\sx{Priors kept on until the end.}}   %
\sx{With the window extended to the end of training (from 5\% to 100\%; the priors are on at 71 of the 75 evaluations), the six at $10^{-3}$ end above the window run at the same rate on dimensions, the torus and the grid in every seed (by at least 2.8, 3.8 and 24.6 points), and on numbers, cities and countries by 0.6, 1.5 and 0.5 points (7, 7 and 6 of 8 seeds). They end at 91.4--99.8\% on all six (Section~\ref{sec:several}), on countries level with the window run at $2\times10^{-3}$ ($+0.3$, $[-0.3,+0.8]$).}   %

\sx{The final validation loss, 1.2112 bits per byte, is 0.1\% above that of the model without priors at its best rate for this loss, which is the lowest rate we ran ($2\times10^{-3}$, 1.2098), 0.4\% above that of its best rate in each seed (95\% interval up to $+0.86\%$), and 0.18\% below that of the window run. These runs are post hoc and ran on A100, H100 and B200 GPUs; re-running a seed on another GPU changes its curve average by about 3 points.}   %

\paragraph{\sx{Larger models.}}   %
\sx{We ran the six priors at 190M ($768\times12$), 505M ($1024\times24$) and 1.07B ($1280\times36$) parameters with the 53M recipe (300M tokens, the same window, sites and weights), changing only the architecture and the learning rate (A100, H100 and B200 GPUs). The six priors, in the window and kept on until the end, and the dimensions prior alone use $6\times10^{-4}$, the rate chosen for the helix alone at these sizes, with seeds 16--18 (in the window also 19). The model without priors runs with seeds 16--18 at five to seven rates from $1.2\times10^{-3}$ to $7.68\times10^{-2}$ (the lowest with seed 16 only), its best rate chosen on the mean over the three seeds, and at $6\times10^{-4}$ with seeds 16--19 to compare the loss at the same rate. On cities and countries, unlike the other four concepts (\S\ref{sec:several}), the runs with and without priors do not separate.}   %

\sx{At every size the six raise the curve average on all six manifolds in every seed, on cities and countries by 4.2 and 3.4 points at 190M, 3.4 and 3.0 at 505M and 2.5 and 2.9 at 1.07B (the other four in \S\ref{sec:several}). They also end higher in every seed on numbers, dimensions and the grid (by 27 to 52 points) and on the torus (by 3 to 6), while on cities and countries the final difference is about zero at 505M and 1.07B. The best rate without priors lies inside the range for every manifold at every size, so these gains are final; at 1.07B it is $3.84\times10^{-2}$ for dimensions and the grid, and at $7.68\times10^{-2}$ the models without priors are worse in all three seeds. Without priors, numbers stays at or below 32.3\% at every size and rate, so the gain on numbers, 47--51 points at every size from 53M to 1.07B, does not depend on which rate is chosen.}   %

\sx{Measured per seed as the first evaluation at which the running mean of three reaches the best final accuracy without priors at any rate, the six need 8--38\% of training on numbers, dimensions, the torus and the grid at 190M (three seeds) and 505M (seed 16), against 10--56\% at 53M by this measure; on cities and countries they need 20--61\% at 190M or do not get there. The joint lift on dimensions over the dimensions prior alone holds at 190M in all three seeds ($+10.4$ points of curve average, $+10.6$ at the end), where the prior alone loses dimensions after its window in two of the three seeds, and, preliminarily, at 1.07B ($+6.6$ averaged over training, positive in two of three seeds; $+5.9$ at the end, positive in all three), but not at 505M ($-1.5$ and $-1.8$). The 1.07B lift rests on three seeds ($t=1.3$--$1.8$), and against all eight seeds of the six priors, unpaired, it is $+1.2$ points averaged over training and $-0.6$ at the end. Without priors, the validation loss is lowest at $6\times10^{-4}$, the lowest rate we ran, so the loss is compared at that rate (\S\ref{sec:several}; $+0.0030$ and $+0.0152$ bits per byte at 190M and 505M). At that rate, across the 15 size $\times$ benchmark comparisons (four paired seeds each, zero-shot, Table~\ref{tab:downstream}), mean differences range from $-0.45$ to $+0.56$ points; none is worse, 9 are non-inferior at a 1-point margin and 6 are undetermined, as with the single number prior at 505M (0 worse, 6 non-inferior and 4 undetermined of 10). LAMBADA perplexity is 3--8\% higher with the six priors (intervals include 1). At 1.07B the undetermined intervals reach down to $-1.9$ to $-2.3$ points, so a small decline is not excluded there, and HellaSwag scores are only 2--4 points above chance.}   %

\sx{At 1.07B the six-prior runs sit on a learning-rate plateau at $6\times10^{-4}$ (one seed): doubling it moves no concept's training-averaged accuracy by more than 3.8 points and keeps the lead over the model without priors at its best rate on all six concepts, while halving it costs 21 points on the grid (84.0 to 62.9\%) and raises the bits per byte from 1.084 to 1.096.}   %

\FloatBarrier   %

\FloatBarrier   %
\section{\nw{Details of the Main Figures and Tables}}
\label{app:capnotes}

\paragraph{\nw{Figure~\ref{fig:curves}.}}
\nw{\og{(a,~e)~Two periodic coordinates of the target helix (period 100 or 10) against the value. (b,~c,~f,~g)~Axes: the period-$P$ component of the means over $x \bmod P$ and \fz{the main direction of the higher place, so the panels show the number-aligned part of the representation and do not establish a helix on their own.} (b,~c)~Layer 4 at the end of training, on the first operand of ``\texttt{Q:~$x$~+~0~=~$x$}'' after web text (rerun on another GPU type). (f,~g)~Layer 16 after 3 epochs of fine-tuning. \fz{(d)~Averaged over the six operand-length combinations; the shuffled helix uses the prior's learning rate and ran on another GPU type. (h)~Ten epochs (the five-fold speed-up in the text is from 30-epoch runs, \S\ref{sec:window}); the shuffled helix has the same strength as the prior.}}} %

\paragraph{\nw{Table~\ref{tab:helped}.}}
\sx{Addition in fine-tuning (note $b$): in 30-epoch runs; the model ends near 95\% without the prior and at 100\% with it.}   %
\nw{Months: among test problems whose reverse was not trained, 39 are answered correctly only with the prior and 9 only without it ($p=1.5\times10^{-5}$, 3 seeds pooled).} %
\og{Months (note $c$): \sx{no held-out (month, answer) pair occurs in training, and} each seed has its own split.}   %
\nw{Weekdays: 5 seeds, $p=0.001$; a target that clusters the answers without the circle gives $+0.9$.} %
\og{Weekdays (note $d$): \sx{the test measures transfer across equivalent shifts, and} the prior acts on the states that predict the answer.}   %
\fz{Vector sums (note $f$): \sx{a one-layer network,} with the prior only at the lowest of three learning rates, against 0 of 5 seeds in each of 28 settings without it (some added post hoc; some diverge or stop updating); one cell of a screen.}   %
\sx{Six priors (note $e$): 8 fresh seeds; the torus and the grid gain mainly while their priors are on; kept on until the end, all six end above the model without priors (\S\ref{sec:several}).}   %

\paragraph{\nw{Figure~\ref{fig:scaling}.}}
\og{\fz{(a)~At 505M, the best learning rates of the grids we tried; 53M and 190M ran on another cluster (same-cluster comparison: Figure~\ref{fig:tpp}b).} \sx{(b)~93M, 136M, 294M, 391M and 731M are the cells of Table~\ref{tab:pt300}; all sizes but 1.07B ran on L40S GPUs.}} %

\paragraph{\nw{Figure~\ref{fig:tpp}.}}
\sx{(a)~At 300M tokens, the solid line joins the eight sizes run on L40S GPUs, the cells of the pre-registered size test, and a dashed segment leads to 1.07B on B200 GPUs; the 294M cell on B200 GPUs ($+29.1$) lies on the L40S one ($+29.0$) and is not drawn (Table~\ref{tab:pt300}). Sizes are total parameters in millions.} %
\sx{(b)~All runs on A100 GPUs of one cluster. At 300M tokens, 53M and 190M were rerun there at the learning rates chosen earlier, and 93M and 294M were added in a pre-registered extension, each arm at the learning rate chosen on seeds 0--3 ($4.8\times10^{-3}$ without the prior at 93M, $9.6\times10^{-3}$ at 294M, and $6\times10^{-4}$ with it at both).} %

\paragraph{\nw{Figure~\ref{fig:pretraining}.}}
\nw{Held-out accuracy is averaged over the six operand-length combinations. (a)~``Step 0'' is a self-timed prior run from the first step, which with 2\% arithmetic stays on for 68--100\% of training on the operands and throughout on web text.} %
\og{Operands are 0--999, and held-out problems are evaluated in training position. (c)~The prior is a window with $\lambda=5$, and the shuffled helix acts on the separate prompts.} %

\paragraph{\nw{Figure~\ref{fig:window}.}}
\nw{(e)~The seven domains are addition, subtraction, number words, Roman numerals, weekdays, months and letters.} %
\nw{In both model families the shuffled helix is below the model without a prior at every checkpoint where it was run.} %

\paragraph{\nw{Table~\ref{tab:costs}.}}
\nw{General ability: \og{three} non-inferior verdicts lie within 0.1 points of the margin, and HellaSwag scores are only 1.5--2.1 points above a randomly initialised model. \fz{The configuration of note $b$ costs 0.058--0.069 bits per byte and is 1.1--3.3 points lower on every benchmark (descriptive).}} %

\end{document}